\documentclass[twoside]{article}

\usepackage[accepted]{aistats2021}

\usepackage[round]{natbib}
\usepackage{graphicx}
\usepackage{caption}
\usepackage{subcaption}
\usepackage{float}
\usepackage{cancel}
\usepackage{multirow}
\usepackage[colorlinks=true,citecolor=blue,linkcolor=blue,allcolors=blue,allbordercolors={1 1 1}]{hyperref}
\usepackage{dblfloatfix}

\usepackage{amsmath}
\usepackage{amsfonts}
\usepackage{bm}
\usepackage{amsthm,amssymb}

\newcommand{\x}{{\mathbf{x}}}
\newcommand{\y}{{\mathbf{y}}}
\newcommand{\Y}{{\mathbf{Y}}}
\newcommand{\z}{{\mathbf{z}}}
\newcommand{\G}{{\mathbf{G}}}

\newcommand{\boldEpsilon}{{\bm{\epsilon}}}

\newcommand{\estimator}[1]{\mathcal{L}_{#1}(\theta)}

\usepackage{soul}

\usepackage[switch]{lineno}

\usepackage{fontawesome}

\begin{document}

\runningtitle{Neural Empirical Bayes}

\twocolumn[

\aistatstitle{Neural Empirical Bayes: Source Distribution Estimation \\and its Applications to Simulation-Based Inference}

\aistatsauthor{ Maxime Vandegar \And Michael Kagan \And Antoine Wehenkel \And Gilles Louppe}

\aistatsaddress{ SLAC National\\ Accelerator Laboratory\\ \And SLAC National\\ Accelerator Laboratory \And ULiège \And ULiège } ]

\begin{abstract}
We revisit empirical Bayes in the absence of a tractable likelihood function, as is typical in scientific domains relying on computer simulations.
We investigate how the empirical Bayesian can make use of neural density estimators first to 
use all noise-corrupted observations to estimate a prior or source distribution over uncorrupted samples, and then to perform single-observation posterior inference using the fitted source distribution. 
We propose an approach based on the direct maximization of the log-marginal likelihood of the observations, examining both biased and de-biased estimators, and comparing to variational approaches. 
We find that, up to symmetries, a neural empirical Bayes approach recovers ground truth source distributions.
With the learned source distribution in hand, we show the applicability to likelihood-free inference and examine the quality of the resulting posterior estimates.
Finally, we demonstrate the applicability of Neural Empirical Bayes on an inverse problem from collider physics. \href{https://github.com/MaximeVandegar/NEB}{\faGithub}
\end{abstract}

\section{Introduction}

The estimation of a \textit{source} distribution over latent random variables $\x$ which give rise to a set of observations $\y$, after undergoing a potentially non-linear corruption process (i.e., a pushforward), is an inverse problem frequently of interest to the scientific and engineering communities. The source distribution, $p(\x)$, may represent the distribution of plausible measurements, or intermediate random variables in a hierarchical model, prior to corruption by a measurement or detection apparatus. The source distribution is of scientific interest as it allows comparison with theoretical predictions and for posterior inference for subsequent observations. 
Notably, in many scientific domains, the relationship between the source and observed distributions is encoded in a simulator
that provides an approximation of the corruption process and generates samples from the likelihood $p(\y|\x)$.
However, as is typical with computer simulations, the likelihood function is implicit and rarely known in a tractable closed form.

Formally, we state the problem of likelihood-free source estimation as follows. 
Given a first dataset $\mathbf{Y}=\{\y_i\}_{i=1}^N$ of $N$ noise-corrupted observations $\y_i$ and a second dataset $\{\x_j, \y_j\}_{j=1}^M$ of matching pairs of source data and  observations, with $\x_j \sim \pi(\x)$ drawn from an arbitrary proposal distribution $\pi(\x)$ and $\y_j \sim p(\y|\x_j)$, our aim is to learn the source distribution $p(\x)$, not necessarily equal to $\pi(\x)$, that has generated the observations $\mathbf{Y}$.
For the class of problems we consider, we may assume that the dataset of $(\x_j, \y_j)$ pairs is generated beforehand using a simulator of the stochastic corruption process.

The source distribution estimation problem is closely related to likelihood-free inference~\citep[LFI,][]{Cranmer2020TheFO}, though there are notable differences in problem statements. 
First, in Bayesian LFI, the objective is the computation of a posterior given a known prior and an implicit likelihood function. In our problem statement, the primary objective is rather to identify an unknown prior or source distribution that, once identified, then enables likelihood-free inference.
Second, we only assume access to a pre-generated dataset of pairs of simulated source data and observations.
In many settings, simulators are highly complex, with long run times to generate data.
As such, sequential methods based on active calls to the simulator,  as often found in the LFI literature, would be impractical.

In this work, we follow an empirical Bayes~\citep[EB,][]{robbins1956, Dempster_1977} approach to address this challenge, using modern neural density estimators to approximate both the intractable likelihood and the unknown source distribution.
Our method, which we call Neural Empirical Bayes (NEB), proceeds in two steps.
First, using simulated pairs $\{\x_j, \y_j\}_{j=1}^{M}$, we use neural density estimation to learn an approximate likelihood. 
Second, by modeling the source distribution with a parameterized generative model, the log-marginal likelihood of the observations is approximated with Monte Carlo integration, and the parameters of the source distribution learned through gradient-based optimization. 
While our estimator of the log-marginal likelihood is biased, it is consistent and the use of deep generative models allows for fast and parallelizable Monte Carlo integration to mitigate its bias. Nonetheless, we also examine de-biased and variational estimators for comparison.
Finally, once a source distribution and likelihood function are learned, we demonstrate that posterior inference for new observations may be performed with suitable sampling-based methods.

We first review EB and describe our NEB approach in Section~\ref{sec:EB}, followed by an examination of log-marginal likelihood estimators in Section~\ref{sec:EB_est}. Related work is discussed in Section~\ref{sec:RW}. 
In Section~\ref{sec:Exp}, we present benchmark problems that explore the efficacy of NEB and provide comparison baselines, as well as a demonstration on a real-world application to collider physics. 
Further discussion and a summary are in Section~\ref{sec:summary}. In addition, we provide a summary of the notations in Appendix \ref{appendix:notationq}.

\section{Empirical Bayes}\label{sec:EB}

Methods for EB~\citep{robbins1956, Dempster_1977} are usually divided into two estimation strategies~\citep{efron2014two}:
either modeling on the $\x$-space, called $g$-modeling; or on the $\y$-space, called $f$-modeling.

Here, we revisit $g$-modeling to learn a source distribution that regenerates the observations $\mathbf{Y}$.
Specifically, we parameterize the source distribution as $q_{\theta}(\x)$ which, when passed through the likelihood $p(\y|\x)$, results in a distribution $q_\theta(\y)$ over noisy observations. 
The log-marginal likelihood of the observations $\mathbf{Y}$ is expressed as
\begin{align}
    \log q_\theta(\Y) &= \sum_{i=1}^N \log q_\theta(\y_i) \nonumber \\
         &= \sum_{i=1}^N \log \int p(\y_i|\x) q_{\theta}(\x)d\x, \label{LogMarginalObjective_1}
\end{align}
and its direct maximization with respect to the parameters $\theta$ leads to a solution for the source distribution. 

The maximization of the log-marginal likelihood is equivalent to the minimization of the Kullback–Leibler divergence $\text{KL}(p(\y)||q_\theta(\y)) = \mathbb{E}_{p(\y)}\left[-\log q_\theta(\y)\right] + \cancel{\text{C}}
                    \approx -\frac{1}{N}\sum_{i=1}^N \log q_\theta(\y_i).$
Therefore, as $\mathbf{Y}$ increases, an optimal solution will correspond to a source distribution that exactly reproduces the observed distribution when passed through the corruption process.
We note however that the maximization of Eq.~\ref{LogMarginalObjective_1} is an ill-posed problem: distinct source distributions may result in the same distribution over observations when folded through the corruption process. As a result, the learned source distribution may differ from the ground truth, for instance missing modes, but still reproduce the observed distribution. We discuss approaches to mitigate these undesired behaviors effects when \textit{a priori} known properties of the source distribution are available.

In the likelihood-free setting, the likelihood function $p(\y|\x)$ is only implicitly defined by the simulator 
which prevents the direct estimation of Eq.~\ref{LogMarginalObjective_1}.  
However, a dataset $\{\x_j,\y_j\}_{j=1}^M$ can be generated beforehand by drawing uncorrupted samples $\x_j$ from a proposal distribution $\pi(\x)$ and running the simulator to generate corresponding noise-corrupted observations $\y_j$. 
Similarly to \citet{Diggle} and \citet{DAGOSTINI1995487} who built likelihood function estimators with kernels or histograms, we use the generated dataset to train a surrogate $q_\phi(\y|\x)$ of the likelihood function, but we make use of modern neural density estimators such as normalizing flows~\citep{tabak2010density, NF}.
After the upfront simulation cost of generating the training data, no additional call to the simulator is needed. 

We optimize the parameters $\phi$  by maximizing the total log-likelihood $\sum_{m=1}^M \log q_{\phi}(\y_m|\x_m)$ with mini-batch stochastic gradient ascent. 
Again, for large $M$, this is equivalent to minimizing $\mathbb{E}_{\pi(\x)}\text{KL}(p(\y|\x)||q_{\phi}(\y|\x))$ and given enough capacity the surrogate likelihood is guaranteed to be a good approximation of $p(\y|\x)$ in the support of the proposal distribution $\pi(\x)$. As a consequence, the support of $\pi(\x)$ should be chosen to cover the full range of plausible source data values.  For example in the simulation-based inference setting, $\pi(\x)$ may be the distribution obtained from a simulator.

\section{Log-marginal likelihood estimation}\label{sec:EB_est}

In Section~\ref{section:biased_estimator} (resp.~\ref{section:unbiased_estimator}) we build a biased (resp. unbiased) estimator of the log-marginal likelihood $\log q_\theta(\y)$ 
with a generative model $\G_{\theta}(\cdot): \mathcal{E} \xrightarrow{} \mathcal{X}$ that defines a differentiable mapping from a base distribution $p(\boldsymbol{\epsilon})$ to $q_{\theta}(\x)$. Then, in Section~\ref{section:VEB}, we show how to use variational estimators of $\log q_\theta(\y)$ for NEB.

\subsection{Biased estimator}
\label{section:biased_estimator}
Given a likelihood function $p(\y|\x)$ or its surrogate $q_{\phi}(\y|\x)$ we define an estimator $\mathcal{L}_K(\theta)$ of the log-marginal likelihood. This estimator can be plugged in Eq.~\ref{LogMarginalObjective_1} to optimize the source distribution parameters $\theta$ by stochastic minibatch gradient ascent. Based on Monte Carlo integration, the estimator is defined as:
    \begin{align}
        \log q_\theta(\y)
        &= \log \mathbb{E}_{q_{\theta}(\x)} \left[p(\y|\x)\right] \nonumber \\
        &= \log \mathbb{E}_{p(\bm{\epsilon})} \left[p(\y|\G_{\theta}(\bm{\epsilon}))\right] \nonumber \\
        & \approx \log \frac{1}{K}\sum_{k=1}^K
     p(\y|\G_{\theta}(\bm{\epsilon}_k)) \nonumber \\
     &= \text{logSumExp} \left[\log p(\y|\G_{\theta}(\bm{\epsilon}_k))\right] - \text{C} \nonumber\\
     &=: \mathcal{L}_K(\theta) \label{estimator},
    \end{align}

where $\bm{\epsilon}_k \sim p(\bm{\epsilon})$, $C$ is a constant independent of $\theta$, and the log-sum-exp trick is used for numerical stability. 
While a large number  $K$ of samples may be needed for good Monte Carlo approximation, this difficulty is alleviated by the ease of generating large samples of source data with the neural sampler $\mathbf{G}_\theta$. 

We study and prove properties of the estimator $\mathcal{L}_K(\theta)$ in Appendix~\ref{appendix:consistency_proof}.
Using the Jensen's inequality, we first show that $\mathcal{L}_K(\theta)$ is biased. 
We demonstrate however that both its bias and its variance decrease at a rate of $\mathcal{O}(\frac{1}{K})$.
Then, similarly to \cite{Burda2016ImportanceWA}, we show
that $\mathcal{L}_K(\theta)$  is monotonically non-decreasing in expectation with respect to $K$, i.e.
\begin{equation}
\label{eq:property_2}
    \mathbb{E}\left[\mathcal{L}_{K+1}(\theta)\right] \ge \mathbb{E}\left[\mathcal{L}_{K}(\theta)\right]. 
\end{equation}
As $K \to \infty$, we finally show that the estimator is however consistent:
\begin{equation}
\label{eq:property_1}
    \lim_{K \xrightarrow{} \infty} \mathcal{L}_K(\theta)
    = \log q_{\theta}(\y). 
\end{equation}

\subsection{Unbiased estimator}
\label{section:unbiased_estimator}

Using the Russian roulette estimator~\citep{RussianRoulette}, we de-bias the log-marginal likelihood estimator $\mathcal{L}_K(\theta)$ as
\begin{equation}
\label{eq:unbiased_estimator}
    \hat{\mathcal{L}}_K(\theta) := \mathcal{L}_K(\theta) + \eta(\theta),
\end{equation}
where $\eta(\theta)$ is a random variable -- whose expectation corrects for the bias -- defined as
\begin{equation}
    \eta(\theta) = \sum_{j=0}^J \frac{\estimator{K+j+1} - \estimator{K+j}}{P(\mathcal{J} \ge j)},
\end{equation}
with $J \sim P(J)$.
Similarly to \cite{luo2020sumo} in their study of Importance Weighted Auto-Encoders \citep[IWAEs,][]{Burda2016ImportanceWA}, we prove in Appendix \ref{appendix:unbiased_proof} that $ \hat{\mathcal{L}}_K $ is an unbiased estimator as long as $P(J)$ is a discrete distribution such that $P(\mathcal{J} \ge j)>0, \forall j>0$. 
Ideally, the distribution $P(J)$ should be chosen such that it adds only a small computational overhead, while providing a finite-variance estimator.
In our experiments, we reduce the computational overhead by re-using the same Monte Carlo terms used for $\mathcal{L}_{K + j}$ to compute $\mathcal{L}_{K + j + 1}$.

\subsection{Variational empirical Bayes}
\label{section:VEB}

For EB in high-dimension, \cite{wang2019} proposed to build upon \cite{DBLP:journals/corr/KingmaW13} and to introduce a variational posterior distribution $q_{\psi}(\x|\y)$ whose parameters $\psi$ are jointly optimized with the parameters $\theta$ of the source distribution by maximizing the evidence lower bound (ELBO):
\begin{align}
       \log q_{\theta}(\y) & \ge \log q_{\theta}(\y) - \text{KL}(q_{\psi}(\x|\y)||p(\x|\y)) \nonumber\\
       \begin{split}&=\mathbb{E}_{q_{\psi}(\x|\y)} \left[ \log p(\y|\x)\right]  \nonumber \\ & \qquad \quad -\text{KL}(q_{\psi}(\x|\y)||q_{\theta}(\x))\end{split}\\
       &=: \mathcal{L}^\text{ELBO} \label{eq:Elbo}.
\end{align}
When $\mathcal{L}^\text{ELBO}$ is optimized with stochastic gradient descent, an unbiased estimator can be obtained with Monte Carlo integration -- usually only one Monte Carlo sample is used which yields a tractable objective. While being tractable, the ELBO is a lower bound (and biased estimator) of the log-marginal likelihood. A common approach \citep{NF} to tighten the bound is to model $q_{\psi}(\x|\y)$ from a large distribution family so that it can closely match the posterior distribution, i.e. efficiently minimize $\text{KL}(q_{\psi}(\x|\y)||p(\x|\y))$.  
Close to our work, IWAEs trade off  computational complexity to obtain a tighter log-likelihood lower bound derived from importance sampling.
Specifically, IWAEs are trained to maximize
\begin{equation}
\label{eq:IWAEs_estimator}
        \mathcal{L}_{K}^\text{IW}(\theta, \psi) = \log \frac{1}{K} \sum_{k=1}^K p(\y|\x_k) w(\x_k),
\end{equation}
where $w(\x_k)=\frac{q_{\theta}(\x_k)}{q_{\psi}(\x_k|\y)}$ and $\mathbf{x}_k \sim q_{\psi}(\x|\y)$. IWAEs are a generalization of the ELBO based on importance weighting  (setting $K=1$ retrieves the ELBO objective). 
\citet{nowozin2018debiasing} showed that the bias and variance of this estimator vanish for $K \xrightarrow{} \infty$ at the same rate $\mathcal{O}(\frac{1}{K})$ as $\mathcal{L}_{K}$.

By design, $\mathcal{L}^\text{ELBO}$ and $\mathcal{L}^\text{IW}$ require the evaluation of the density of new data points under the source model, whereas $\mathcal{L}_\text{K}$ and $\hat{\mathcal{L}}_K$ only require the efficient sampling from the source model.
Which method to use should therefore depend on the downstream usage of the source distribution.
While the evaluation of densities required by $\mathcal{L}^\text{ELBO}$ and $\mathcal{L}^\text{IW}$ limits the range of models that can be used and makes the introduction of inductive bias more difficult, $\mathcal{L}_\text{K}$ and $\hat{\mathcal{L}}_K$ can be used with any generative model. In the rest of the manuscript we refer indistinguishably to the source distribution as $q_{\theta}(\x)$ although the generative models used with  $\mathcal{L}_\text{K}$ and $\hat{\mathcal{L}}_K$ may not allow its evaluation. In that case, we mean the pushforward distribution implied by $\G_{\theta}(\x)$.

\section{Related work}\label{sec:RW}

\paragraph{Empirical Bayes} 
In the most common forms of $g$-modeling, the likelihood function and the prior distribution are chosen such that the marginal likelihood can be computed and maximized iteratively or analytically. More recent approaches model the prior distribution analytically but assume both the $\x-$space and $\y-$space are finite and discrete~\citep{Narasimhan2016AGP, article_EB_Efron}. Then, given a known likelihood function encoded in tensor form, the distribution parameters are optimized by maximum marginal likelihood estimation. Similarly to this latter approach, we do not require a likelihood function in closed-form, but we build a continuous surrogate that allows its direct evaluation rather than discretizing it.  

While \citet{wang2019} only theoretically proposed using Eq.~\ref{eq:Elbo} in EB, we show experimentally in the next section the applicability of this method. 
Concurrent work~\citep{dockhorn2020density} also used this approach to solve a density deconvolution task on Gaussian noise processes. 
Our work differs as we show the applicability of these methods on much more complicated black-box simulators, including a real inverse problem from collider physics. 
Black-box simulators imply that a neural network surrogate replaces the likelihood function, and thus, learning $\theta$ and $\psi$ requires to backpropagate through the surrogate. 

Finally, in the context of likelihood-free inference, \cite{Louppe2019AdversarialVO} used adversarial training for learning a prior distribution such that, when corrupted by a non-differentiable black-box model, reproduces the empirical distribution of the observations. This can be seen as  $g$-modeling EB where a prior distribution is optimized based on observations.

\paragraph{Unfolding} 
Approximating a source distribution $p(\x)$ given corrupted observations is often referred to as unfolding in the  particle physics literature~\citep[for reviews see][]{Cowan:2002in,Blobel:2203257,Adye:2011gm}. A common approach \citep{Richardson, Lucy_deconvolution, DAGOSTINI1995487} is to discretize the problem and replace the integral in Eq.~\ref{LogMarginalObjective_1} with a sum, resulting in a discrete linear inverse problem. The surrogate model $q_{\phi}(\y|\x)$ of the likelihood function is encoded in tensor form while $q_{\theta}(\x)$ is modeled with a histogram. These approaches are typically limited to low dimensions. 
In order to scale to higher dimensions,  preliminary work by \cite{neural_unfolding} explored the idea of modelling $q_{\phi}(\y|\x)$ and $q_{\theta}(\x)$ with normalizing flows to approximate the integral in Eq.~\ref{LogMarginalObjective_1} with Monte Carlo integration. 
Aiming to the same objective, \cite{Andreassen2019OmniFoldAM} replaced the sum in discrete space with a full-space integral using the likelihood ratio which is used for re-weighting. \cite{bellagente2020invertible} used invertible neural networks for learning a posterior that can be used for unfolding while our EB approach focuses on learning a source distribution at inference time.

\paragraph{Likelihood-free inference} 
The use of a surrogate model of the likelihood function that enables inference as if the likelihood was known is not new. 
Since \cite{Diggle}, kernels and histograms have been vastly used for 1D density estimation. 
More recently, several Bayesian likelihood-free inference algorithms~\citep{Papamakarios2019SequentialNL, SNPEA, SNPEB, APT, hermans2019likelihoodfree, Durkan2020OnCL} have been developed to carry out inference when the likelihood function is implicit and intractable. 
These methods all operate by learning parts of the Bayes' rule, such as the likelihood function, the likelihood-to-evidence ratio, or the posterior itself, and all require the explicit specification of a prior distribution.
By contrast, the primary objective of our work is to learn a prior distribution from a set of noise-corrupted observations which, once it is identified, then enables any of the aforementioned Bayesian LFI algorithms for posterior inference.
We refer the reader to \cite{Cranmer2020TheFO} for a broader review of likelihood-free inference.

\begin{table}[!h]
\centering
\setlength{\tabcolsep}{2pt}
\renewcommand{\arraystretch}{1.5}
\scriptsize
\begin{tabular}{llll}
\hline
\hline
                 Simulator           & K & $\mathcal{L}_K$ & $\hat{\mathcal{L}}_K$ \\ \hline
\multirow{4}{*}{SLCP}       &  10  & $0.82_{\pm0.01}$ &  $0.65_{\pm0.04}$  \\ 
                            &  128 & $0.57_{\pm0.01}$& $0.59_{\pm0.02}$ \\  
                            &  256 & $0.55_{\pm0.02}$ & $0.54_{\pm 0.00}$  \\  
                            &  1024 & $0.53_{\pm0.01}$ & $0.52_{\pm 0.01}$  \\ \hline
\multirow{4}{*}{Two-moons}  &  10  &  $0.69_{\pm0.02}$ &   $0.56_{\pm0.02}$   \\  
                            &  128 &    $0.53_{\pm0.01}$ & $0.57_{\pm 0.06}$  \\ 
                            &  256 & $0.52_{\pm0.01}$ &  $0.52_{\pm0.02}$     \\ 
                            &  1024 &  $0.52_{\pm 0.01}$ & $0.53_{\pm0.01}$    \\ \hline
\multirow{4}{*}{IK}  & 
                              10 & $0.80_{\pm0.13}$ &   $0.67_{\pm0.08}$   \\ 
                     &  128 &   $0.65_{\pm 0.04}$ &  $0.67_{\pm0.12}$    \\ 
                  &  256 &  $0.66_{\pm 0.02}$ &  $0.71_{\pm0.09}$     \\ 
                            &  1024 &  $0.66_{\pm 0.03}$ &  $0.62_{\pm0.03}$  \\  \hline
\hline
\end{tabular}
\caption{\label{table:biased_vs_unbiased}
ROC AUC between the observed distribution $p(\mathbf{y})$ and the regenerated distribution $\int p(\mathbf{y}|\mathbf{x})q_{\theta}(\mathbf{x})d\mathbf{x}$.
The closer to $0.5$, the better the estimation in $\y$-space.
\textit{NEB successfully identifies source distributions that result in distributions over noise-corrupted observations that are almost indistinguishable from the ground truth.
When $K$ is low, de-biasing leads to substantial improvements.}}
\end{table}
\begin{table*}[hbt!]
\centering
\setlength{\tabcolsep}{3pt}
\renewcommand{\arraystretch}{1.5}
\scriptsize
\begin{tabular}{l|lll|lll|lll}
\hline\hline
\multirow{2}{*}{Simulator}                                   & \multicolumn{3}{c|}{$\x$-space}                                        &
\multicolumn{3}{c|}{$\y$-space}    &   \multicolumn{3}{c}{$\x$-space (symmetric prior)}                                                                       \\ 
                                                             & $\mathcal{L}^{\text{ELBO}}$ & $\mathcal{L}_{128}^\text{IW}$ & $\mathcal{L}_{1024}$ &  $\mathcal{L}^{\text{ELBO}}$ & $\mathcal{L}_{128}^\text{IW}$ & $\mathcal{L}_{1024}$ & $\mathcal{L}^{\text{ELBO}}$ & $\mathcal{L}_{128}^\text{IW}$ & $\mathcal{L}_{1024}$ \\ \hline
SLCP  &  $1.00_{\pm 0.00}$ & $0.82_{\pm 0.09}$ & $0.75_{\pm 0.03}$ &   $0.92_{\pm0.04}$ &     $0.50_{\pm 0.00}$  &    $0.53_{\pm 0.01}$ & $0.99_{\pm0.01}$  &  $0.59_{\pm0.05}$ & $0.81_{\pm0.02}$ \\ 
Two-Moons & $0.75_{\pm 0.00}$ & $0.75_{\pm 0.00}$ & $0.55_{\pm 0.02}$  &          $0.50_{\pm 0.01}$  & $0.50_{\pm 0.00}$ & $0.52_{\pm 0.01}$ & $0.51_{\pm0.01}$ &  $0.50_{\pm0.01}$ & $0.51_{\pm0.02}$\\ 
\begin{tabular}[c]{@{}l@{}}IK\end{tabular} & $1.00_{\pm 0.00}$ & $0.95_{\pm0.05}$ & $0.74_{\pm 0.03}$  & $0.51_{\pm 0.01}$     &    $0.50_{\pm 0.01}$              &    $0.62_{\pm 0.03}$      & $0.97_{\pm0.01}$ &   $0.72_{\pm0.02}$ & $0.66_{\pm0.04}$               \\ \hline\hline
\end{tabular}
\caption{\label{tab:ROAC_AUC_major_table}Source estimation for the benchmark problems.
ROC AUC between $q_{\theta}(\x)$ and $p(\x)$ ($\x$-space), and between the observed distribution $p(\mathbf{y})$ and the regenerated distribution $\int p(\mathbf{y}|\mathbf{x})q_{\theta}(\mathbf{x})d\mathbf{x}$ ($\y$-space).}
\end{table*}
\section{Experiments}\label{sec:Exp}
We present three studies of NEB. In Section~\ref{sec:source_estimation}, we analyze the intrinsic quality of the recovered source distribution for the estimators discussed in Section~\ref{sec:EB_est}. In Section~\ref{sec:posterior_inference}, we explore posterior inference with the learned source distribution. Finally, in Section~\ref{section:DetectorEffectCorrection} we show the applicability of NEB on an inverse problem from collider physics. All experiments are repeated 5 times with $q_{\phi}(\y|\x)$ and $q_{\theta}(\x)$ relearned in each experiment. Means and standard deviations are reported.

\subsection{Source estimation}
\label{sec:source_estimation}
We evaluate NEB on three benchmark problems: (a) a toy model
with a simple likelihood but complex posterior (SLCP) introduced by \cite{Papamakarios2019SequentialNL}, (b) the two-moons model of \cite{APT}, and (c) an inverse kinematics problem (IK) proposed by \cite{analyzingINNs}. 
See Appendix~\ref{appendix:simulators} for complementary experimental details.
We use datasets of $M=15000$ samples to train  surrogate models $q_{\phi}(\y|\x)$ for each simulator.
All density models are parameterized with normalizing flows made of four coupling layers \citep{dinh2014nice, Dinh2017DensityEU}.
Further architecture and optimization details can be found in Appendix~\ref{appendix:hyper_params}. 
The source distributions $q_\theta(\x)$ are optimized on $N=10000$ observations $\y$ and we show further results with only one or two observations in Appendix~\ref{appendix:M_1_EB}.
The ground truth source distributions $p(\x)$ are $\mathcal{U}(-3, 3)^5$ for SLCP, $\mathcal{U}(-1, 1)^2$  for two-moons and $\mathcal{N}(\mathbf{0}, \mathbf{Diag}(\frac{1}{4}, \frac{1}{2}, \frac{1}{2}, \frac{1}{2}))$ for IK.

\paragraph{Biased vs. unbiased estimator} 

We first compare the biased and unbiased estimators $\mathcal{L}_K$ and $\hat{\mathcal{L}}_K$. 
Table~\ref{table:biased_vs_unbiased} reports the ROC AUC scores of a classifier trained to distinguish between noise-corrupted observations from the ground truth $p(\y)$ and noise-corrupted observations from the marginal $\int p(\y|\x) q_{\theta}(\x)d\x$ obtained by passing source data from $q_{\theta}(\x)$ into the exact simulator.
For both estimators, the table shows that we successfully identify a source distribution $q_{\theta}(\x)$ resulting in a distribution over noise-corrupted observations which is almost indistinguishable from the ground truth $p(\y)$.
When $K$ is low, de-biasing the estimator leads to significant improvements. 
When $K$ increases, the bias of $\mathcal{L}_K$ drops quickly, and de-biasing, which introduces variance, does not significantly improve the results. 
Therefore, we recommend using the de-biased estimator when $K$ is constrained to be low, e.g., when the GPU memory is limited. 
In the following, we set $K=1024$ and only consider the biased estimator $\estimator{K}$.

\paragraph{Monte Carlo vs. variational methods}
We evaluate the quality of $\mathcal{L}_K$, $\mathcal{L}^{\text{ELBO}}$ and $\mathcal{L}_{K}^\text{IW}$. 
For a fair comparison, we use an Unconstrained Monotonic Neural Network autoregressive flow  \citep[UMNN-MAF,][]{Wehenkel2019UnconstrainedMN} to parameterize the prior for all losses. 
The recognition network $q_{\psi}(\x|\y)$ for the variational approaches is modeled with the same architecture as the prior, but is conditioned on $\y$. 
We use $K=128$ for $\mathcal{L}_{K}^\text{IW}$ due to GPU memory constraints. We show in Appendix~\ref{appendix:MLP_as_prior} that simpler implicit generative models can be used with the Monte Carlo estimators $\mathcal{L}_K$ and $\hat{\mathcal{L}}_K$, effectively reducing inference time and allowing to use higher values of $K$.

\begin{figure}[h]
    \centering
    \includegraphics[width=.9\linewidth]{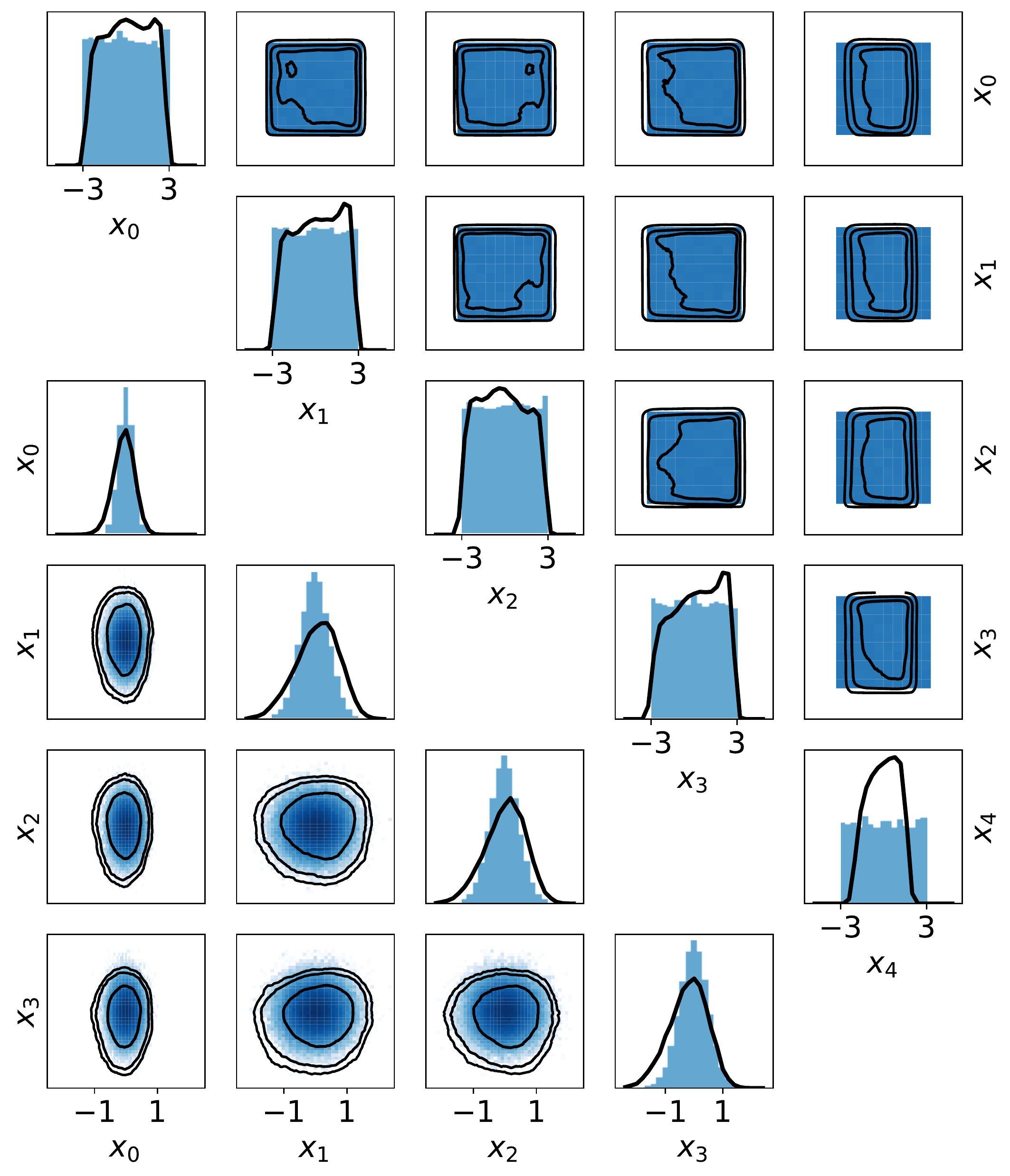}
    \caption{\label{fig:LearnedPrior}
    Source estimation results for $\mathcal{L}_{1024}$ on SLCP (top) and IK (bottom). 
    The source distribution $p(\x)$ are shown in blue against the estimated source distribution $q_\theta(\x)$ in black (the 68-95-99.7\% contours are shown).
\textit{The identified source distributions are similar to the unseen source distributions.}}
\end{figure}

\begin{figure}
    \centering
    \includegraphics[width=.72\linewidth]{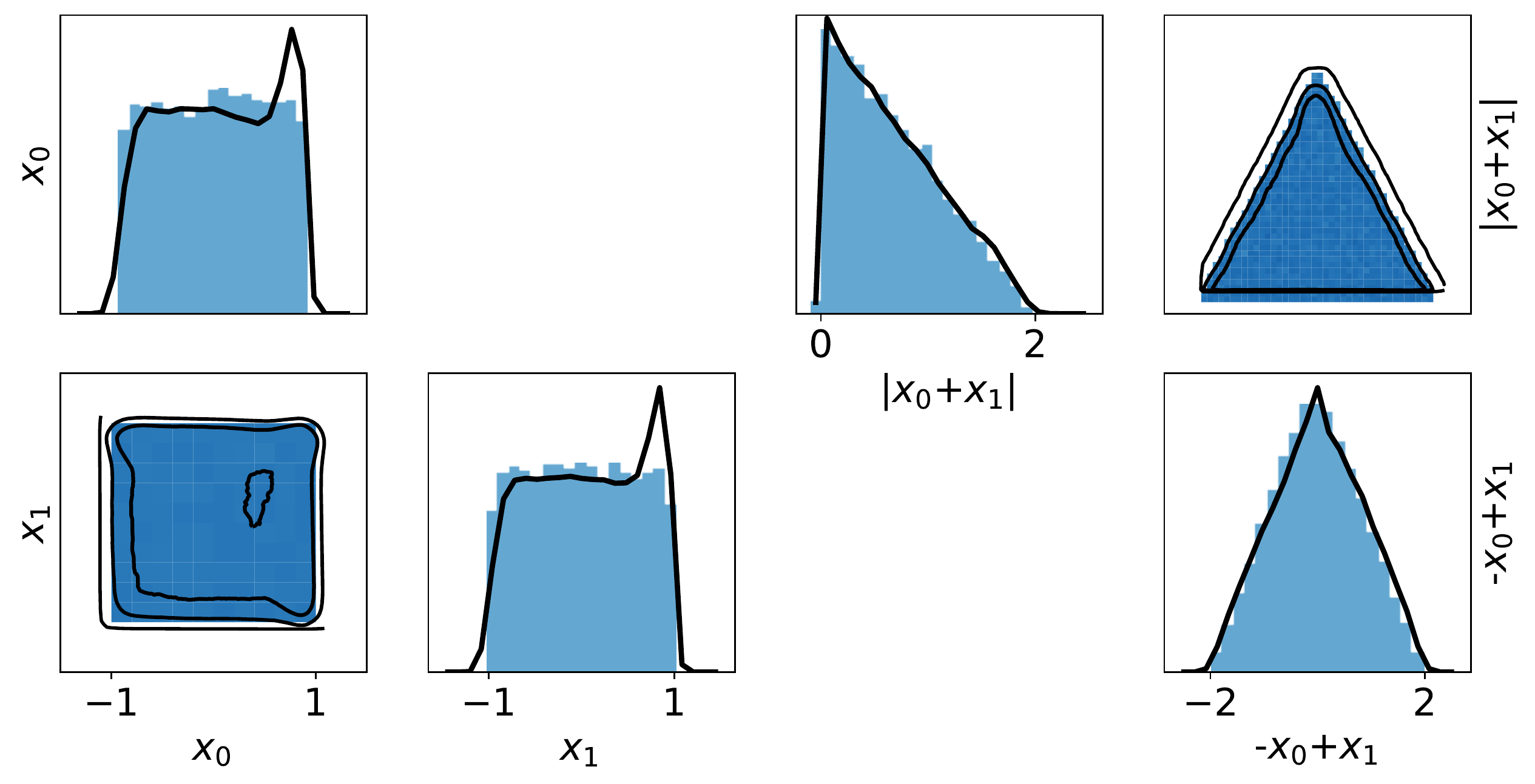}
    \caption{\label{fig:LearnedPrior_2dMoons} 
    Source estimation results for $\mathcal{L}_{1024}$ on the two-moons problem. 
    The source distribution $p(\x)$ is shown in blue against the estimated source distribution $q_\theta(\x)$ in black (the 68-95-99.7\% contours are shown).
    \textit{As shown on the right, up to the symmetries of the problem, the identified source distribution matches the unseen source distribution.}}
\end{figure}

Table~\ref{tab:ROAC_AUC_major_table} shows the ROC AUC of a classifier trained to discriminate between samples from the ground truth source distribution $p(\x)$ and samples from the source distribution $q_{\theta}(\x)$ identified by each of the different methods. 
A ROC AUC score between $0.5$  and  $0.7$  is often considered poor discriminative performance, therefore indicating good source estimation. 
The estimator $\mathcal{L}_{1024}$ leads to the most accurate source distributions on these three tasks. In particular, the source distribution found for the two-moons problem is almost perfect. At the same time, the results for SLCP and IK are marginally acceptable, and largely better than for the variational methods ($\mathcal{L}_K^{\text{IW}}$ and $\mathcal{L}^{\text{ELBO}}$). 
Figures \ref{fig:LearnedPrior} and \ref{fig:LearnedPrior_2dMoons} illustrate for $\mathcal{L}_{1024}$ how the exact and learned sources distributions are visually similar.

Table~\ref{tab:ROAC_AUC_major_table} also reports the discrepancy between the corrupted data from the identified source distributions and the ground truth distribution of noise-corrupted observations. 
While $\mathcal{L}^{\text{ELBO}}$ does not give good results for SLCP, tightening the evidence lower-bound with $\mathcal{L}^{\text{IW}}_{128}$ yields good results on all problems. 
While $\mathcal{L}_{1024}$ has similar performance to $\mathcal{L}^{\text{IW}}_{128}$ on SLCP and two-moons, it is performing worse for IK, due to the difficulty of approximating $\int p(\y|\x)p(\x)d\x$ from Monte Carlo integration since the likelihood function for this problem is almost a Dirac function (see Appendix \ref{appendix:simulators} for more details).

After observing the different estimators' reconstruction quality, the superiority of $\mathcal{L}_{1024}$ on source estimation over variational methods may be surprising at first glance. 
However, the variational methods require learning both a source distribution and a recognition network that are consistent with the likelihood function and the observations. 
This means that a wrong recognition network may prevent learning the correct source distribution as they must be consistent with each other. 
In the three experiments analyzed here, the prior distribution is a simple unimodal continuous distribution, whereas the posteriors are discontinuous and multimodal. In these cases, learning only the source distribution is simpler than learning both a source and posterior distributions. 

\paragraph{Symmetric source distribution}
As mentioned before, multiple optimal solutions may co-exist when the inverse problem is ill-posed. 
On close inspection, figures~\ref{fig:LearnedPrior} and \ref{fig:LearnedPrior_2dMoons} show that NEB successfully recovers the domain of the source data but fails to exactly reproduce the ground truth source distribution. 
Indeed, for all problems considered here, the passage of $\x$ through the corruption process results in a loss of information in $\y$, which may lead to multiple solutions.
We observe this in Figure~\ref{fig:LearnedPrior_2dMoons}, where we plot the quantities $|\x_1+\x_2|$ and $-\x_1 + \x_2$ that are sufficient statistics of $\x$ for estimating $\y$.
We see that the distribution over these intermediate variables is nearly equal for the ground truth distribution and the identified source distribution. 
This indicates that, up to symmetries, NEB recovers the ground truth source distribution.

\begin{figure}
    \centering
    \includegraphics[width=0.32\textwidth]{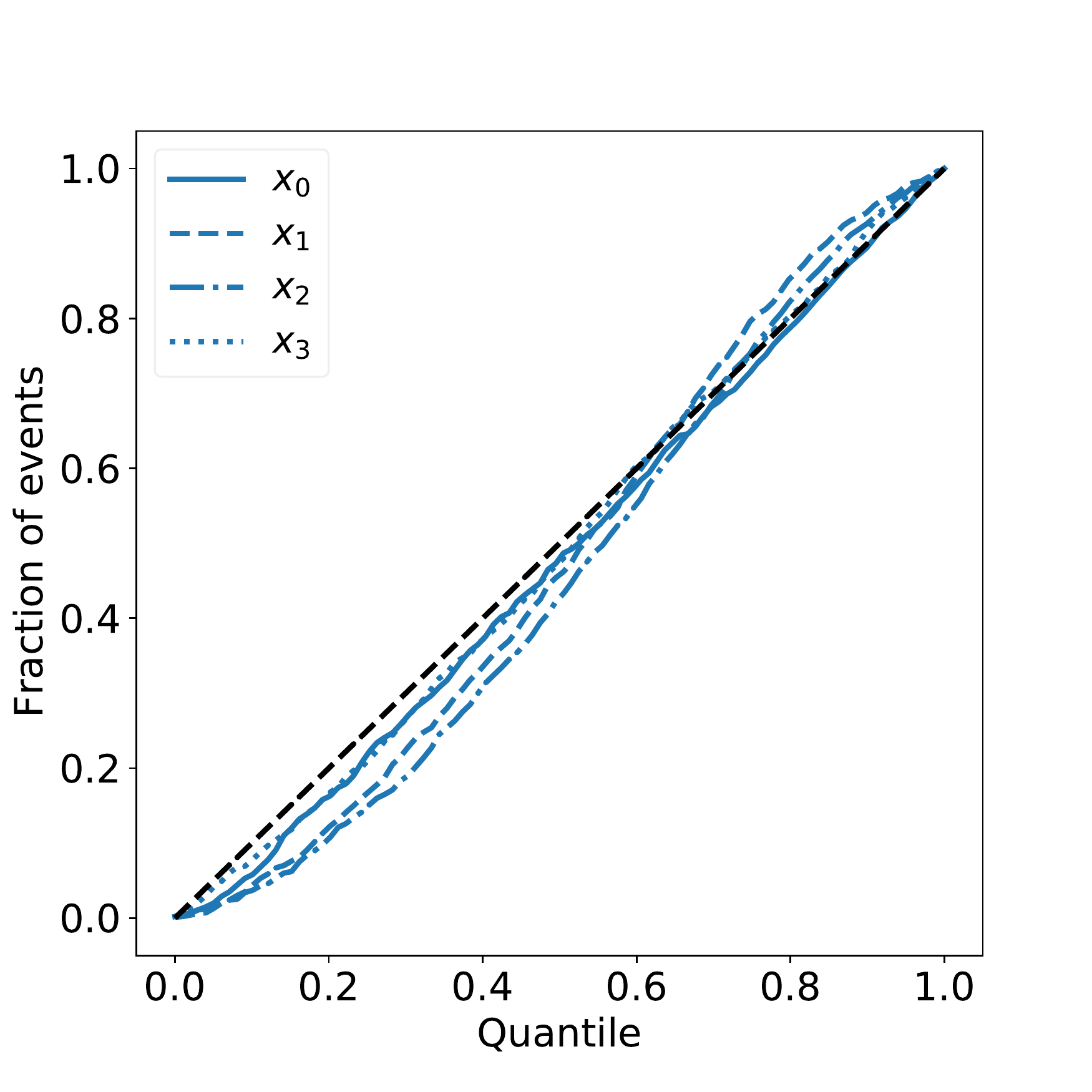}
    \caption{\label{fig:KS_calibration}
    Posterior inference for IK. 
    The plot shows the per-parameter calibration curves obtained with rejection sampling using the learned likelihood $q_{\psi}(\y|\x)$ and the identified source distribution $q_{\theta}(\x)$ with $\mathcal{L}_{1024}$. 
    \textit{The curves indicate a reasonably well calibrated posterior distribution.}}
\end{figure}

A reasonable way to encourage learning a good source distribution is to enforce \textit{a priori} known properties such as its domain, symmetries, or smoothness. 
This type of useful inductive biases can be embedded in the neural network to constrain the solution space. As such, we modify the UMMN-MAF 
networks $q_{\theta}(\x)$ so that the generated distributions are one-to-one symmetric, i.e. $q_\theta(\left[x_1, ..., x_d\right]) = q_\theta(\left[\pm x_1, ..., \pm x_d\right]$) (these modifications are detailed in Appendix \ref{appendix:symmetric_UMNN_MAF}). 
Table \ref{tab:ROAC_AUC_major_table} shows that the symmetric distributions are more similar to the unseen distributions $p(\x)$ in all but one case. 
For example, all methods learn to approximately identify the exact source distribution on the two-moons despite the simulator's destructive process. 
Results for $\mathcal{L}_{1024}$ on SLCP are worse because the regularization pushes the learned distribution to a solution that still reproduces the observed distribution with high accuracy (the ROC AUC between the observed distribution and the regenerated one drops to $0.51_{\pm0.01}$), but that moves away from the unseen source distribution.
Further inductive bias should therefore be introduced; 
for example, the learned distribution can be bounded using specific activation functions in the last layer of $\G_{\theta}(\cdot)$.
We note here that the non-variational methods are generally better suited for inductive bias as their architecture designs are less constrained.

\subsection{Likelihood-free posterior inference}
\label{sec:posterior_inference}
In the context of Bayesian posterior inference, the source distribution we retrieve with NEB can used as a prior distribution.
Therefore, the learned prior $q_{\theta}(\x)$ together with the surrogate likelihood $q_{\phi}(\y|\x)$ unlock the subsequent likelihood-free estimation of the posterior $p(\x|\y)$ -- for which the fidelity will depend on both the 
correctness of the source distribution and the likelihood. There is an ongoing debate in the EB literature regarding the use of the data twice for posterior inference in this approach \citep{gelman2008, Darnieder2011BayesianMF, Gelman_2017}. When few prior knowledge are available, EB allows to learn insights from the data, and we believe it is valuable to incorporate those insights in the prior rather than, for instance, choosing a wide prior and especially in high dimensions where the prior choice is important \citep{Gelman_2017}.


As we have used normalizing flows with $\mathcal{L}_K$ and $\hat{\mathcal{L}}_K$, state-of-the-art Markov Chain Monte Carlo (MCMC) methods such as Hybrid Monte Carlo (HMC) can be used for sampling the posterior. Other generative models that do not allow density evaluation could be used in our empirical Bayes setup but would not permit the usage of MCMC. In this section, we focus on rejection sampling, rather than MCMC, as the source distribution model allows fast parallel sampling which makes the algorithm efficient even when the acceptance rate is low.
We perform rejection sampling as follows: given $u \sim \mathcal{U}(0, 1)$, we accept samples $\x \sim q_{\theta}(\x)$ such that $u < \frac{q_{\phi}(\y|\x)}{M}$ where $M>q_{\phi}(\y|\x), \forall \x$ is determined empirically. 
On the other hand, variational approaches ($\mathcal{L}^{\text{ELBO}}$ and $\mathcal{L}^{\text{IW}}_{K}$) directly learn a posterior $q_\psi(\x|\y)$ as a function of the single observation $\y$, which enables immediate per-event posterior inference. 
For $\mathcal{L}^{\text{IW}}_{K}$, \cite{cremer2017reinterpreting} suggest using importance sampling.


\begin{figure}
    \centering
    \includegraphics[width=.9\linewidth]{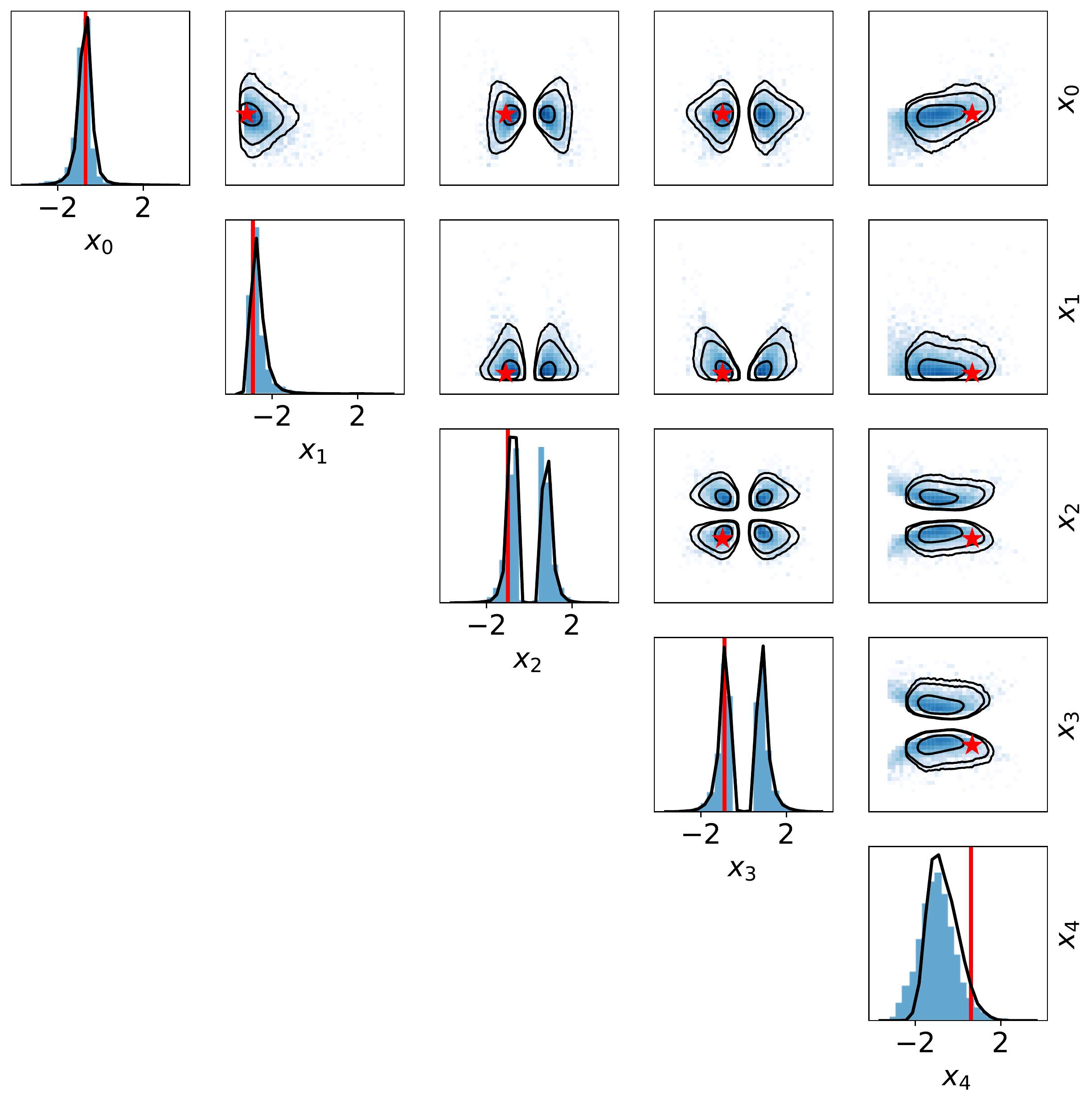}
    \caption{\label{fig:comparison_with_sequential_methods}Posterior distribution obtained from MCMC with the exact source distribution and the exact likelihood function on  SLCP in blue against the posterior distribution obtained with $q_{\phi}(\y|\x)$ and $q_{\theta}(\x)$ learned from $\mathcal{L}_{1024}$ in black (the 68-95-99.7\% contours are shown). Generating source sample $\x$ are indicated in red.
    \textit{The approximated posterior distribution closely matches the ground truth.}}
    \label{fig:my_label}
\end{figure}

We assess the goodness of the posterior distributions with a calibration test. Inspired by \cite{bellagente2020invertible}, for multiple observations $\y_i$, we approximate the 1D posterior distributions and report the fraction of events as a function of the quantile to which the generating source data $\x_i$ fall. 
Figure~\ref{fig:KS_calibration} reports calibration curves associated with $\mathcal{L}_{1024}$ for the IK problem, indicting 
reasonably well calibrated posteriors. We should note however that this observation is a necessary but not sufficient condition for well-calibrated posterior distributions. More details and quantitative results are given in Appendix~\ref{appendix:sbc}.

\begin{figure*}[ht]
\begin{subfigure}{.33\textwidth}
\captionsetup{width=0.95\textwidth}
  \centering
  \includegraphics[width=1.\linewidth]{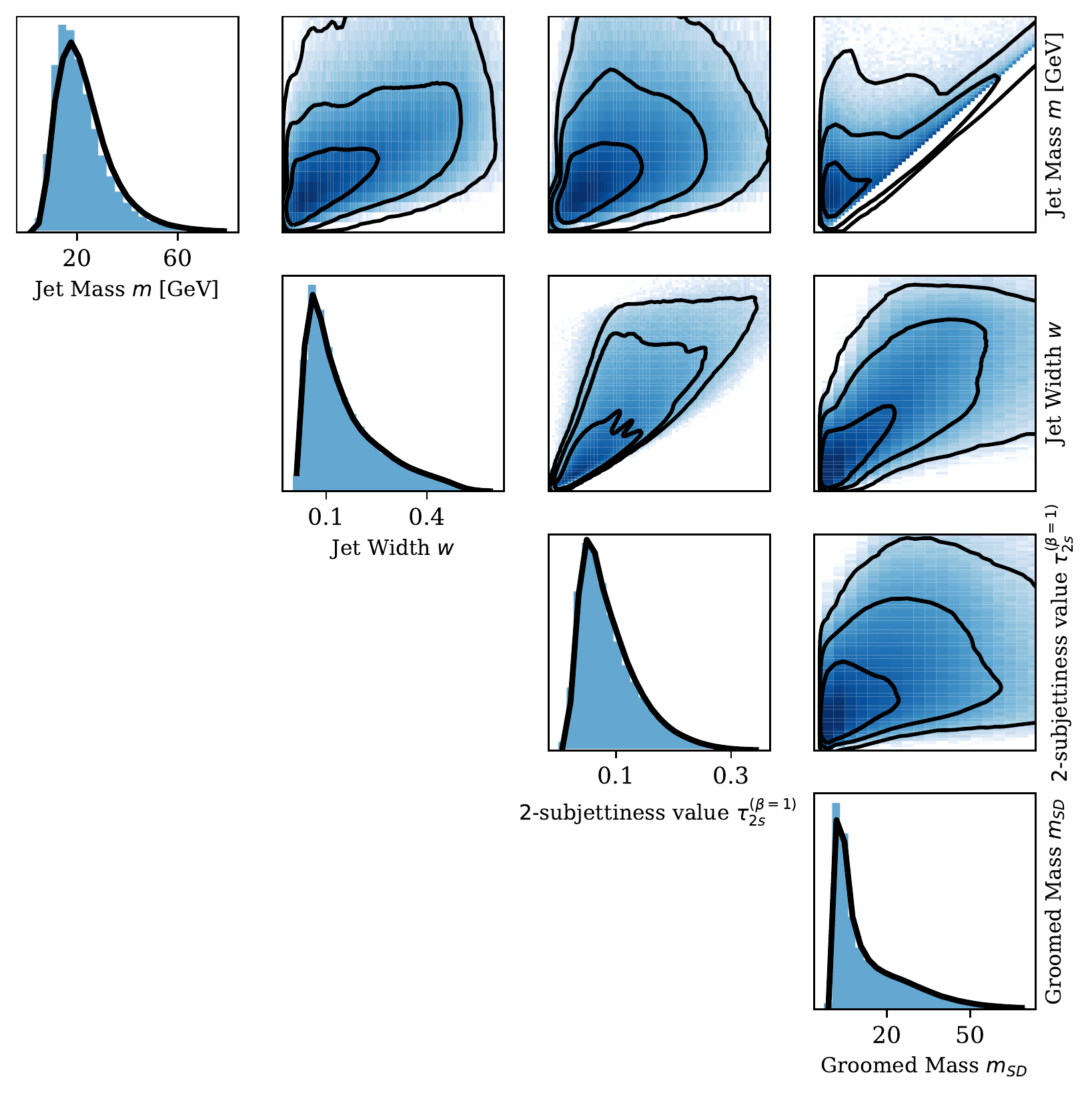}
  \caption{\label{fig:Herwig_learned_prior}}
\end{subfigure}
\begin{subfigure}{.33\textwidth}
\captionsetup{width=0.95\textwidth}
  \centering
  \includegraphics[width=1.\linewidth]{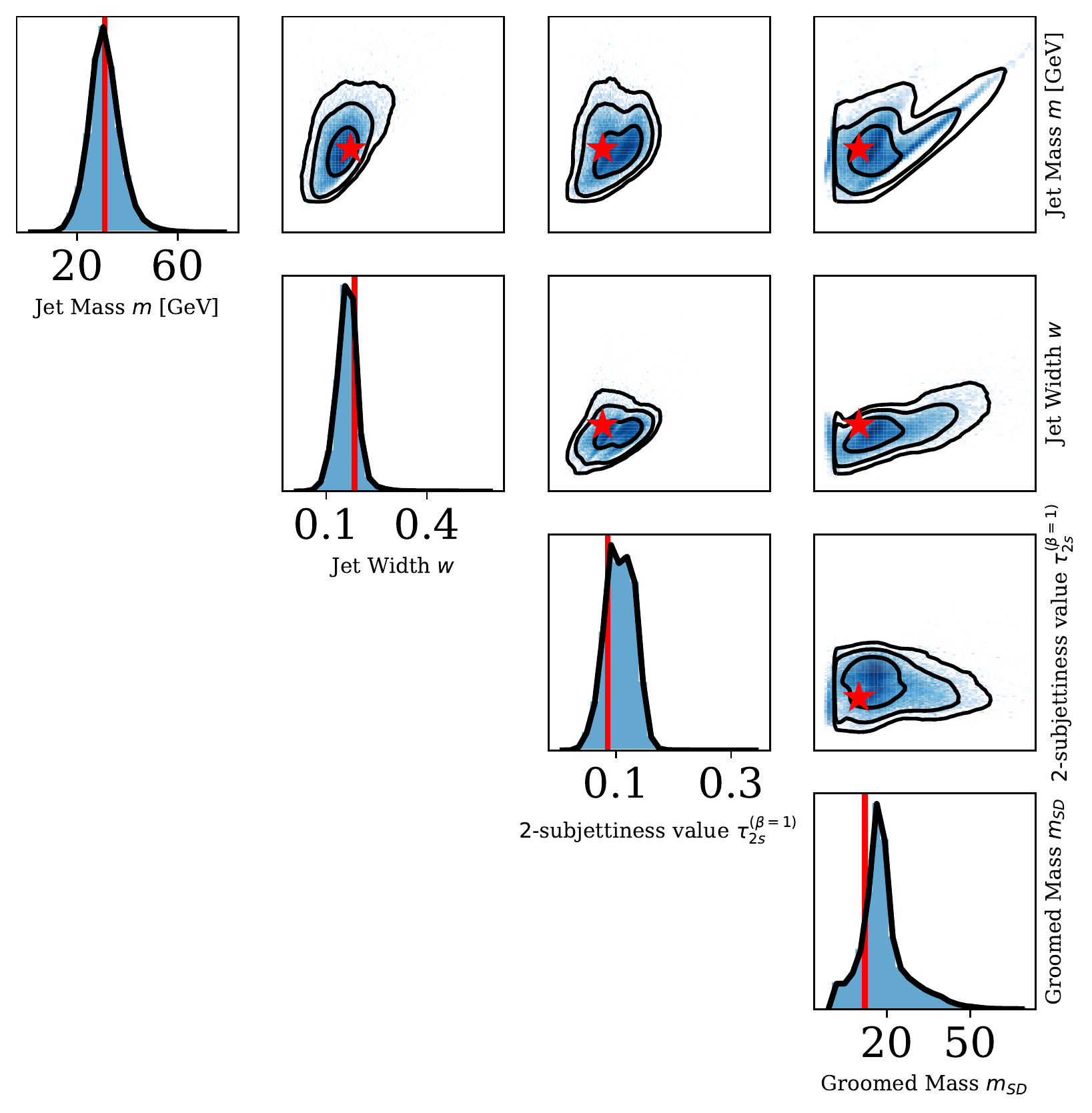}  
  \caption{\label{Posterior_distribution_physics}}
\end{subfigure}
\begin{subfigure}{.33\textwidth}
\captionsetup{width=0.95\textwidth}
  \centering
  \vspace{0.1cm}
  \includegraphics[width=1.\linewidth]{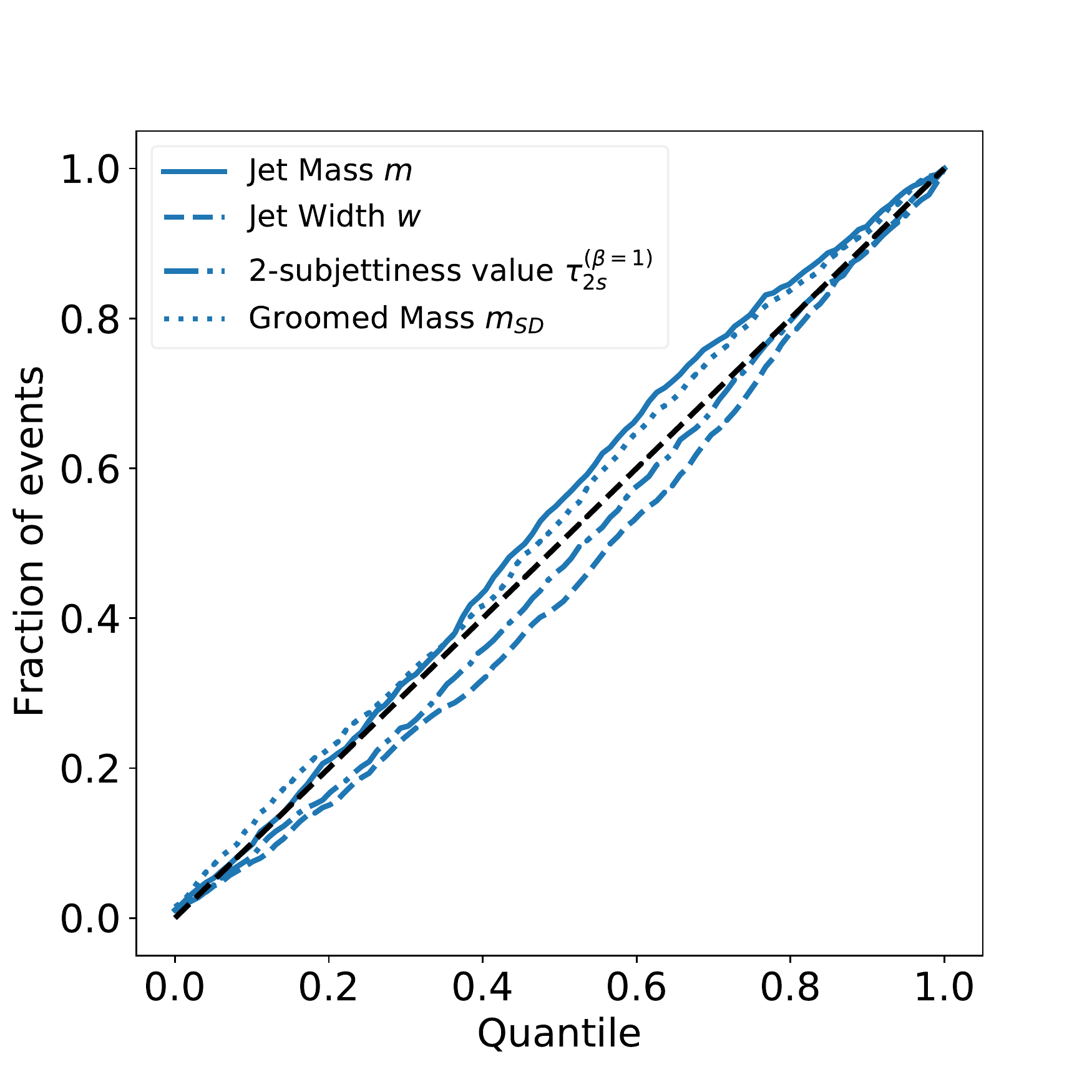}
  \vspace{-0.4cm}
  \caption{\label{Calibration_test_herwig}}
\end{subfigure}
\caption{Neural Empirical Bayes for detector correction in collider physics.
(a) The source distribution $p(\x)$ is shown in blue against the estimated source distribution $q_\theta(\x)$ in black.
(b) Posterior distribution obtained with rejection sampling, with generating source sample $\x$ indicated in red.
(c) Calibration curves for each jet property obtained with rejection sampling on $10000$ observations. In (a) and (b), contours represent the 68-95-99.7\% levels.
}

\end{figure*}

Finally, we show in Figure \ref{fig:comparison_with_sequential_methods} an example of posterior distribution obtained with rejection sampling using $q_{\theta}(\x)$ and $q_{\phi}(\y|\x)$ as learned with $\mathcal{L}_{1024}$, against the ground truth posterior obtained with Markov Chain Monte Carlo using the exact likelihood and source data distribution. We emphasize that NEB can recover nearly the exact posterior with no access to the likelihood function or to the prior distribution. To the best of our knowledge, this is the first work to show posterior inference is possible in this extreme setting.

\subsection{Detector correction in collider physics}
\label{section:DetectorEffectCorrection}

At colliders like the LHC, the distribution of particles produced from an interaction and incident on detectors can be predicted from theoretical models. Thus measurements of such distributions can be used to directly test theoretical predictions. However, while detectors measure the energy and momentum of particles, they also induce noise due to the stochastic nature of particle-material interactions and of the signal acquisition process. Thus a key challenge in comparing measurements to theoretical predictions is to correct noisy detector observations to obtain experimentally observed incident particle source distributions.  This is frequently done by binning 1D or 2D distributions and solving a discrete linear inverse problem. Instead we apply NEB for estimating the multi-dimensional source distribution. We use the publicly available simulated dataset~\citep{andreassen_anders_2019_3548091} of paired source and corrupted measurements of properties of jets, or collimated streams of particles produce by high energy quarks and gluons. Simulation details are found in Appendix~\ref{appendix:phys_simulator}.

Surrogate training was performed using one source simulator as a proposal distribution. The same surrogate architecture and hyperparameters as in the toy experiments were used (see Appendix~\ref{appendix:hyper_params} for details). We assess NEB in a dataset with the source distribution produced by a different simulator of the same physical process. Both datasets for surrogate training and source distribution learning contain approximately 1.6 million events. This is an example setting where sequential inference methods cannot be used as only a fixed dataset is available and not the simulator.

\begin{table}[h!]
 \centering
 \setlength{\tabcolsep}{2pt}
\renewcommand{\arraystretch}{1.5}
\scriptsize
 \begin{tabular}{llll} \hline  \hline    & $\mathcal{L}^\text{ELBO}$ & $\mathcal{L}_{128}^\text{IW}$ & $\mathcal{L}_{1024}$ \\ 
 \hline  $\x$-space & $0.99_{\pm 0.02}$   & $0.63_{\pm 0.06}$ & $0.57_{\pm 0.05}$ \\ 
    $\y$-space  & $0.87_{\pm 0.08}$   & $0.51_{\pm 0.01}$  & $0.50_{\pm 0.01}$\\ 
 \hline \hline   \end{tabular} 
 \caption{Source estimation in collider physics. ROC AUC between $q_{\theta}(\x)$ and the unseen source distribution $p(\x)$ ($\x$-space), and between the observed distribution $p(\mathbf{y})$ and the regenerated distribution $\int q_{\phi}(\mathbf{y}|\mathbf{x})q_{\theta}(\mathbf{x})d\mathbf{x}$.\label{PHYSICS_ROCAUC}}
\end{table}
 
\paragraph{Source estimation} We focus on the $\mathcal{L}_{1024}$ estimator for source distribution learning although we also report results with the other estimators. Optimization is done with Adam using default parameters and an initial learning rate of $10^{-4}$. We train for 10 epochs with minibatches of size 256. The density estimator for the source distribution comprised 6 coupling layers, with 3-layer MLPs with 32 units per layer and ReLU activations used for the scaling and translation functions.
Parameters were determined with a hyperparameter grid search using a held out validation set from the dataset on which the surrogate is trained. The learned source distribution is observed to closely match the true simulated source distribution, as seen in Figure~\ref{fig:Herwig_learned_prior}. Table~\ref{PHYSICS_ROCAUC} reports the ROC AUC between the learned and ground truth distributions, indicating only small discrepancies between them.

\paragraph{Likelihood-free posterior inference} Figure \ref{Posterior_distribution_physics} shows the learned posterior distribution against the generating source data. Plots are scaled to the prior-space. The model learns nicely a region of plausible values for the generating source data.  To assess the quality of the posterior inference on more data, Figure \ref{Calibration_test_herwig} shows the fraction of events as a function of the quantile to which the generating source data belongs under the learned posterior distribution. 
Results indicate reasonably well calibrated posterior distributions.

\section{Summary and discussion}\label{sec:summary}

In this work, we revisit $g$-modeling empirical Bayes with neural networks to estimate source distributions from non-linearly corrupted observations. We propose both a biased and de-biased estimator of the log-marginal likelihood, and examine variational methods for this challenge. 
We show that we can successfully recover source distributions from corrupted observations.    
We find that inductive bias is highly beneficial for solving ill-posed inverse problems and can be embedded in the structure of the neural networks used to model the source distribution.
Although the explored approaches are general, we specifically study the likelihood-free setting, and we successfully perform posterior inference without direct access to either a likelihood function or a prior distribution.

\paragraph{Future work}  
In this work we have mainly examined low-dimensional settings. 
We believe that further analysis of these methods for high-dimensional data such as images and time series could be of strong interest from both a theoretical and practical point of view. In particular, assessing the computational challenges of each method and the importance of inductive bias in this challenging setting are promising directions towards improvements in solving high-dimensional inverse problems.

\subsubsection*{Acknowledgments}
We thank Johann Brehmer and Kyle Cranmer for their helpful feedback on the manuscript. We also thank the anonymous reviewers for their thoughtful comments.
Maxime Vandegar and Michael Kagan are supported by the US Department of Energy (DOE) under grant DE-AC02-76SF00515, and Michael Kagan is also supported by the SLAC Panofsky Fellowship. Antoine Wehenkel is a research fellow of the F.R.S.-FNRS (Belgium) and acknowledges its financial support. Gilles Louppe is recipient of the ULiège - NRB Chair on Big data and is
thankful for the support of NRB.

\bibliography{bibliography}

\newpage

\onecolumn

\appendix

\section{ 
\label{appendix:notationq}  Summary of the notations used in the paper}

All notations used in the paper are summarized in Table \ref{tab:notation_recap}.

\begin{table}[H]
\centering
\begin{tabular}{ll}
\hline
\hline
Notation & Definition\\ 
\hline
 $p(\y|\x)$ & Likelihood function implicitly defined by the simulator \\ 
 $q_{\phi}(\y|\x)$ & Surrogate model of $p(\y|\x)$  \\ 
   $p(\y)$ & Observed distribution \\
 $q_{\theta}(\y)$ & Estimator of $p(\y)$  \\ 
 $p(\x)$ & Unseen source distribution that has generated  $p(\y)$\\ 
 $q_{\theta}(\x)$ & Surrogate model of $p(\x)$ \\ 
 $q_{\phi}(\x|\y)$ & Variational posterior distribution \\ 
 $\pi(\x)$ & Proposal distribution used to generate a dataset in \\ 
 &  order to train $q_{\phi}(\y|\x)$ \\ 

 \hline
 \hline
\end{tabular}
\caption{\label{tab:notation_recap} Summary of the notations used in the paper.}
\end{table}

\section{ 
\label{appendix:consistency_proof} Properties of the log-marginal estimators $\mathcal{L}_K$ and $\hat{\mathcal{L}}_K$}
\subsection{Bias of $\estimator{K}$}

The bias of $\mathcal{L}_K$ is derived from the Jensen's inequality:

\begin{subequations}
    \begin{align}
        \mathbb{E}\left[\mathcal{L}_K\right] =& \mathbb{E}\left[\frac{1}{K}\sum_{k=1}^{K} \log p(\y|\G_{\theta}(\boldEpsilon_k)) \right]\\
        =& \frac{1}{K}\sum_{k=1}^{K} \mathbb{E}\left[\log p(\y|\G_{\theta}(\boldEpsilon_k)) \right]\\
        \label{eq:prove_bias_jensens_step}
        \le& \frac{1}{K}\sum_{k=1}^{K} \log \mathbb{E}\left[p(\y|\G_{\theta}(\boldEpsilon_k)) \right]\\
        =& \log q_{\theta}(\y)
    \end{align}
\end{subequations}
where, since the logarithm is strictly concave, the equality in Eq.~\ref{eq:prove_bias_jensens_step} holds iif the random variable $p(\y|\x_k), \text{ } \x_k=\G_{\theta}(\boldEpsilon_k)$ is degenerate, that is $\exists! \mathbf{c}: p(\x_k)=\delta_\mathbf{c}(\x_k)$, which is not the case in general. 

\subsection{Convergence rate of $\estimator{K}$}
Closely following \cite{nowozin2018debiasing}, we show that the bias of the estimator $\estimator{K}$ decreases at a rate $\mathcal{O}(\frac{1}{K})$, in particular:
$$ \mathbb{E}\left[\estimator{K}\right] = \log p(\y) - \frac{1}{K} \frac{\mu_2}{2\mu^2} + \mathcal{O}(\frac{1}{K}),$$
which implies 
$$\mathbb{E}\left[\estimator{K}\right] = \log p(\y) + \mathcal{O}(\frac{1}{K}) .$$

\begin{proof}
Let $w:=p(\y|\x), \x \sim q_\theta(\x)$ and $Y_K := \frac{1}{K} \sum^K_{i=1} w_i$. We have $\gamma := \mathbb{E}\left[Y_K\right] = \mathbb{E}\left[w\right] =: \mu$ because the expectation is a linear operator. Let us expand $\log Y_K$ around $\mathbb{E}\left[w\right]$ with a Taylor series: 
$$\log Y_K = \log \mathbb{E}\left[w\right] - \sum^{\infty}_{j=1}\frac{(-1)^j}{j \mathbb{E}\left[w\right]^j} (Y_K - \mathbb{E}\left[w\right])^j.$$
Taking the expectation with respect to the samples $\x_i$ leads to:
$$\mathbb{E}\left[\log Y_K \right] = \log \mathbb{E}\left[w\right] - \sum^{\infty}_{j=1}\frac{(-1)^j}{j \mathbb{E}\left[w\right]^j} \mathbb{E}\left[(Y_K - \mathbb{E}\left[w\right])^j \right].$$

We can relate the moments $\gamma_i := \mathbb{E}\left[(Y_K - \mathbb{E}\left[Y_K \right])^i \right]$ of the sample mean $Y_K$ to the moments $\mu_i := \mathbb{E}\left[(w - \mathbb{E}\left[w \right])^i \right]$ of the samples $w$ using the Theorem 1 of \citep{angelova_sample_moments}:
\begin{align*}
    \gamma_2 &= \frac{\mu_2}{K}\\
    \gamma_3 &= \frac{\mu_3}{K^2}.
\end{align*}
Expanding the Taylor series to order 3 leads to:
$$\mathbb{E}\left[\log Y_K \right] = \log \mathbb{E}\left[w\right] - \frac{1}{2\mu^2}\frac{\mu_2}{K} + \frac{1}{3\mu^3}\left(\frac{\mu_3}{K^2}\right) + o(\frac{1}{K}),$$
which implies
$$ \mathbb{E}\left[\estimator{K}\right] = \log p(\y) - \frac{1}{K} \frac{\mu_2}{2\mu^2} + \mathcal{O}(\frac{1}{K}).$$
\end{proof}

Again, we directly copy \cite{nowozin2018debiasing} to show the convergence rate of the variance of $\estimator{K}$ to $0$ in $\mathcal{O}(\frac{1}{K})$.
\begin{proof}
Using the definition of the variance and the Taylor series of the logarithm, we have:
\begin{align*}
    \mathbb{V}\left[\log Y_K \right] &= \mathbb{E}\left[(\log Y_K - \mathbb{E}\left[\log Y_K \right])^2\right]\\
    &= \mathbb{E}\left[ \left( \log \mu - \sum^{\infty}_{i=1} \frac{(-1)^i}{i\mu^i}(Y_K - \mu)^i - \log \mu + \sum^{\infty}_{i=1} \frac{(-1)^i}{i\mu^i}\mathbb{E}\left[(Y_K - \mu)^i\right]\right)^2\right]\\
    &= \mathbb{E}\left[ \left(\sum^{\infty}_{i=1}\frac{(-1)^i}{i\mu^i}(\mathbb{E}\left[(Y_K - \mu)^i\right] - (Y_K - \mu)^i\right)^2\right].
\end{align*}
If we expand the last expression to the third order and substitute the samples moments $\gamma_i$ with the central moments $\mu_i$ we eventually obtain:
$$\mathbb{V}\left[\log Y_K \right] = \frac{1}{K}\frac{\mu_2}{\mu^2} - \frac{1}{K^2} (\frac{\mu_3}{\mu^3} - \frac{5\mu^2_2}{2\mu^4}) + o(\frac{1}{K^2}).$$
\end{proof}

\subsection{$\mathcal{L}_K$ non-decreasing with $K$}
Closely following \cite{Burda2016ImportanceWA}, we show the estimator is non-decreasing with $K$, $$\mathbb{E}\left[\mathcal{L}_{K+1}(\theta)\right] \ge \mathbb{E}\left[\mathcal{L}_{K}(\theta)\right].$$
\begin{proof}
Let $I = \{i_1, ..., i_K\} \subset \{1, ..., K + 1\}$ with $|I| = K$ be a uniformly distributed subset of $K$ distinct indices from $\{1, ..., K + 1\}$. We notice that $\mathbb{E}_{I} \left[\frac{\sum_{k=1}^K a_{i_k}}{K}\right] = \frac{\sum_{k=1}^{K + 1} a_k}{K+1}$ for any sequence of numbers $a_1, ..., a_{K+1}$. 

Using this observation and Jensen's inequality leads to
\begin{subequations}

\begin{align}
    \mathbb{E}_{\boldEpsilon_1, ..., \boldEpsilon_{K +1}}\left[\mathcal{L}_{K+1}(\theta)\right] &= \mathbb{E}_{\boldEpsilon_1, ..., \boldEpsilon_{K + 1}}\left[\log \frac{1}{K+1}\sum^{K+1}_{k=1}p(\y|\G_\theta(\boldEpsilon_k))\right]\\
    &=\mathbb{E}_{\boldEpsilon_1, ..., \boldEpsilon_{K + 1}}\left[\log\mathbb{E}_{I} \left[ \frac{1}{K}\sum^{K}_{k=1}p(\y|\G_\theta(\boldEpsilon_{i_k}))\right]\right]\\
    &\ge \mathbb{E}_{\boldEpsilon_1, ..., \boldEpsilon_{K + 1}}\left[\mathbb{E}_{I} \left[\log\frac{1}{K}\sum^{K}_{k=1}p(\y|\G_\theta(\boldEpsilon_{i_k}))\right]\right]\\
    &= \mathbb{E}_{\boldEpsilon_1, ..., \boldEpsilon_{K}}\left[\log\frac{1}{K}\sum^{K}_{k=1}p(\y|\G_\theta(\boldEpsilon_{k}))\right]\\
    &= \mathbb{E}_{\boldEpsilon_1, ..., \boldEpsilon_{K}}\left[{\mathcal{L}_K}\right]
\end{align}
\end{subequations}

\end{proof}

\subsection{$\mathcal{L}_K$ consistency}
We show the consistency of the estimator $\mathcal{L}_K$, that is:

\begin{equation}
    \lim_{K \xrightarrow{} \infty} \mathcal{L}_K(\theta)
    = \log q_{\theta}(\y).
\end{equation}
\begin{proof}
Using the strong law of large numbers:
\begin{subequations}
   \begin{align}
   \label{eq:consistency_proof_estimator definition}
       \lim_{K \xrightarrow{} \infty} \mathcal{L}_K(\theta) &= \lim_{K \xrightarrow{} \infty} \log \frac{1}{K} \sum_{k=1}^K p(\y|\G_{\theta}(\boldEpsilon_k)) \\
     \label{eq:consistency_proof_interchanging_limits}
       &= \log \lim_{K \xrightarrow{} \infty} \frac{1}{K} \sum_{k=1}^K p(\y|\G_{\theta}(\boldEpsilon_k))\\
       \label{eq:consistency_proof_law_large_numbers}
       &= \log \mathbb{E}_{p(\boldEpsilon)} p(\y|\G_{\theta}(\boldEpsilon))\\
       \label{eq:consistency_proof_LOTUS}
       &= \log \mathbb{E}_{q_{\theta}(\x)} p(\y|\x)\\
       \label{eq:consistency_proof_marginalization}
       &= \log q_{\theta}(\y).
   \end{align}
\end{subequations}
In Eq.~\ref{eq:consistency_proof_estimator definition}, we rewrite the definition of the estimator and then, in Eq.~\ref{eq:consistency_proof_interchanging_limits} we interchange the limit and logarithm operators by continuity of the logarithm. In Eq.~\ref{eq:consistency_proof_law_large_numbers}, we use the strong law of large numbers and then, in Eq.~\ref{eq:consistency_proof_LOTUS} we use the LOTUS theorem to rewrite the expectation with respect to the distribution $q_{\theta}(\x)$ implicitly defined by the generative model $\G_{\theta}(\cdot)$. Finally, Eq.~\ref{eq:consistency_proof_marginalization} is obtained by marginalization. 
\end{proof}

\subsection{Unbiased estimator $\hat{\mathcal{L}}_K$} \label{appendix:unbiased_proof}
We want to show this estimator is unbiased, $$\mathbb{E}_{J\sim P(J), \boldEpsilon_1, ..., \boldEpsilon_{K + J} \sim p(\boldEpsilon)}\left[\hat{\mathcal{L}}_K \right] = \log q_{\theta}(\y),$$
where $$\hat{\mathcal{L}}_K = \mathcal{L}_K + \eta$$ with $\eta = \sum_{j=0}^J \frac{\estimator{K+j+1} - \estimator{K+j}}{P(\mathcal{J} \ge j)}$. 
\begin{proof}
Following closely \cite{luo2020sumo}, we proceed as follows. First we observe that:

\begin{subequations}

\begin{align}
    \mathbb{E}_{J\sim P(J), \boldEpsilon_1, ..., \boldEpsilon_{K + J} \sim p(\boldEpsilon)}\left[\hat{\mathcal{L}}_K \right] &= \mathbb{E}_{J\sim P(J), \boldEpsilon_1, ..., \boldEpsilon_{K + J} \sim p(\boldEpsilon)}\left[\mathcal{L}_K + \eta\right]\\
    &= \mathbb{E}_{\boldEpsilon_1, ..., \boldEpsilon_{K} \sim p(\boldEpsilon)}\left[\mathcal{L}_K\right] + \mathbb{E}_{J\sim P(J), \boldEpsilon_1, ..., \boldEpsilon_{K + J} \sim p(\boldEpsilon)}\left[\eta\right],
\end{align}
where we have:
\begin{align}
    \mathbb{E}_{J\sim P(J), \boldEpsilon_1, ..., \boldEpsilon_{K + J} \sim p(\boldEpsilon)}\left[\eta\right] &=
    \mathbb{E}_{J\sim P(J), \boldEpsilon_1, ..., \boldEpsilon_{K + J} \sim p(\boldEpsilon)}\left[\sum_{j=0}^J \frac{\estimator{K+j+1} - \estimator{K+j}}{P(\mathcal{J} \ge j)}\right]\\
    &=\mathbb{E}_{J\sim P(J)}\left[\sum_{j=0}^J  \frac{\mathbb{E}_{\boldEpsilon_1, ..., \boldEpsilon_{K + j + 1} \sim p(\boldEpsilon)}\left[\estimator{K+j+1}\right] - \mathbb{E}_{\boldEpsilon_1, ..., \boldEpsilon_{K + j} \sim p(\boldEpsilon)}\left[\estimator{K+j}\right]}{P(\mathcal{J} \ge j)}\right]\\
    &= \sum_{j=0}^{\infty}\mathbb{E}_{\boldEpsilon_1, ..., \boldEpsilon_{K + j + 1} \sim p(\boldEpsilon)}\left[\estimator{K+j+1}\right] - \mathbb{E}_{\boldEpsilon_1, ..., \boldEpsilon_{K + j} \sim p(\boldEpsilon)}\left[\estimator{K+j}\right] \label{eq:russian_roulette_step}\\ 
    &=\lim_{j \xrightarrow{} \infty} \mathbb{E}_{\boldEpsilon_1, ..., \boldEpsilon_{j} \sim p(\boldEpsilon)}\left[\mathcal{L}_j(\theta)\right] - \mathbb{E}_{\boldEpsilon_1, ..., \boldEpsilon_{K} \sim p(\boldEpsilon)}\left[\estimator{K} \right]\\
    &=\log q_{\theta}(\y) - \mathbb{E}_{\boldEpsilon_1, ..., \boldEpsilon_{K} \sim p(\boldEpsilon)}\left[\estimator{K} \right],
\end{align}
\end{subequations}

where Eq.~\ref{eq:russian_roulette_step} is a property of the Russian roulette estimator (see Lemma 3 of \citep{chen2019residual}) that holds if (i) $P(\mathcal{J} \geq k) > 0, \forall k > 0 $ and (ii) the series converge absolutely. The first condition is ensured by the choice of $P(J)$ and the second condition is also ensured thanks to the non-decreasing and consistency properties of the biased estimator.
\end{proof}

\section{ 
\label{appendix:simulators}  Benchmark problems}

Beyond doing inference on a real simulator from collider physics, we show the applicability of the methods on three benchmark simulators inspired from the literature that are described below. 

\subsection{Simple likelihood and complex posterior (SLCP)}

Given parameters $\x \in \mathbb{R}^5$, the SLCP simulator \citep{Papamakarios2019SequentialNL} generates $\y \in \mathbb{R}^8$ according to:
\begin{subequations}
\label{eq_SLCP}
    \begin{align}
        & \boldsymbol{\mu} = [x_1, x_2]^\top \\
        \label{sim:square_one}
         & s_1 = x_3^2\\
         \label{sim:square_two}
         & s_2 = x_4^2\\
         & \rho = \tanh(x_5)\\
         & \mathbf{\Sigma} = \begin{bmatrix}
            s_1^2 & \rho s_1 s_2\\
            \rho s_1 s_2 & s_2^2
            \end{bmatrix}\\
         & \mathbf{y}_j \sim \mathcal{N}(\boldsymbol{\mu}, \mathbf{\Sigma}), \quad j = 1, . . . , 4\\
         & \mathbf{y} = [\mathbf{y}_1^\top, . . . , \mathbf{y}_4^\top]^\top.
    \end{align}
\end{subequations}

The source data $p(\x)$ is uniform between $[-3, 3]$ for each $x_i$.

\subsection{Two-moons}
 Given parameters $\x \in \mathbb{R}^2$, the the two-moons simulator \citep{DBLP:journals/corr/abs-1907-02392} generates $\y \in \mathbb{R}^2$ according to:
\begin{subequations}
\label{2dMoon_intro_prob}
    \begin{align}
         & a \sim \mathcal{U}(-\frac{\pi}{2}, \frac{\pi}{2}) \\
         & r \sim \mathcal{N}(0.1, 0.01^2)\\
         & \mathbf{p} = [r \cos(a) + 0.25, r \sin(a)]^\top\\
         \label{eq_64_c}
         & \y = \mathbf{p} + [-\frac{|x_1 + x_2|}{\sqrt{2}}, \frac{-x_1 + x_2}{\sqrt{2}}]^\top.
    \end{align}
\end{subequations} 
The source data $p(\x)$ is uniform between $[-1, 1]$ for each $x_i$.

\subsection{Inverse Kinematics} 

\cite{analyzingINNs} introduced a problem where $\x \in \mathbb{R}^4$ but that can still be easily visualized in 2-D. They model an articulated arm that can move vertically along a rail and that can rotate at three joints. Given parameters $\x$, the arm's end point $\y \in \mathbb{R}^2$ is defined as:
\begin{subequations}
    \label{eq:IK}
    \begin{align}
        y_1 &= x_1 + l_1 \sin(x_2) + l_2 \sin(x_2 + x_3) + l_3 \sin(x_2 + x_3 + x_4)\\
        y_2 &= l_1 \cos(x_2) + l_2 \cos(x_2 + x_3) + l_3 \cos(x_2 + x_3 + x_4)
    \end{align}
\end{subequations}
with arm lengths $l_1 = l_2 = 0.5, l_3 = 1.0$.

As the forward model defined in Eq.~\ref{eq:IK} is deterministic and that we are interested in stochastic simulators, we add noise at each rotating joint. Noise is sampled from a normal distribution $\epsilon \sim \mathcal{N}(0, \sigma^2)$ with $\sigma = 0.00017 \text{ rad} \equiv 0.01^{\circ}$.

The source data $p(\x)$ follows a gaussian $\mathcal{N}(0, \sigma_i^2)$ for each $\x_i$ with $\sigma_1 = 0.25 \text{ rad} \equiv 14.33^{\circ}$ and $\sigma_2 = \sigma_3 = \sigma_4 = 0.5 \text{ rad} \equiv 28.65^{\circ}$.

\section{ 
\label{appendix:hyper_params}  Benchmark problems - hyperparameters}

The surrogate models $q_{\phi}(\y|\x)$ are modeled with coupling layers \citep{dinh2014nice, Dinh2017DensityEU} where the scaling and translation networks are modeled with MLPs with ReLu activations. In \cite{Dinh2017DensityEU}, the scaling function is squashed by a hyperbolic tangent function multiplied by a trainable parameter.  We rather use soft clamping of scale coefficients as introduced in \cite{DBLP:journals/corr/abs-1907-02392}:

\begin{equation}
    s_{clamp} = \frac{2\alpha}{\pi}\arctan(\frac{s}{\alpha})
\end{equation}
 which gives $s_{clamp} \approx s$ for $s \ll |\alpha|$ and $s_{clamp} \approx \pm \alpha$ for $|s| \gg \alpha$. We performed a grid search over the surrogate model hyperparameters and found $\alpha = 1.9$ to be a good value for most architectures, as in \cite{DBLP:journals/corr/abs-1907-02392}. Therefore, we fixed $\alpha$ to $1.9$ in all models.

The surrogate models are trained for 300 epochs over the whole dataset of pairs of source and corrupted data. Conditioning is done by concatenating the conditioning variables $\x$ on the inputs of the scaling and translation networks. More details are given in Table \ref{Surrogate:params}.

\begin{table}[H]
\centering
\begin{tabular}{ll}
\hline
\hline
     \multicolumn{2}{l}{\hspace{1.9cm} Architecture} \\ \hline
     Network architecture      & Coupling layers          \\ 
Scaling network       &         3 $\times$ 50   (MLP)  \\ 
Translation network &  3 $\times$ 50    (MLP)        \\ 
N$^\circ$flows             &   4           \\ 
Batch size          &       128       \\
Optimizer          &        Adam      \\ 
Weight decay   &     $5 \times 10^{-5}$         \\ 
Learning rate   & $10^{-4}$             \\ \hline \hline
\end{tabular}
\caption{\label{Surrogate:params}Hyperparameters used to train and model $q_{\phi}(\y|\x)$}
\end{table}

The source data distributions $q_{\theta}(\x)$ are modeled with UMNN-MAFs \citep{Wehenkel2019UnconstrainedMN}. The forward evaluation of these models defines a bijective and differentiable mapping from a distribution to another one which allows to compute the jacobian of the transformation in $\mathcal{O}(d)$ where $d$ is the dimension of the distributions. However, inverting the model requires to solve a root finding algorithm which is not trivially differentiable.  For $\mathcal{L}_K$ and $\hat{\mathcal{L}}_K$, the forward model defines a differentiable mapping from noise $\z$ to $\x$. This design allows to sample new data points in a differentiable way and to evaluate their densities.

For $\mathcal{L}^{\text{ELBO}}$ and $\mathcal{L}^{\text{IW}}_K$, the forward model defines a differentiable mapping from $\x$ to $\z$ which allows to evaluate in a differentiable way the density of any data point $\x$, as required by the two losses. $\mathcal{L}^{\text{ELBO}}$ and $\mathcal{L}^{\text{IW}}_K$ also require to introduce a recognition network $q_{\psi}(\x|\y)$ which should allow to differentially sample new data points and evaluate their densities. Therefore, the same architecture as $q_{\theta}(\x)$ is used. The core architecture of all models is the same and detailed in Table \ref{tab:prior_params}.

$\mathcal{L}_K$ and $\hat{\mathcal{L}}_K$ are trained over 100 epochs over the whole observed dataset. For $\mathcal{L}_K$, $10\%$ of the data were held out to stop training if the loss did not improve for 10 epochs. The $\hat{\mathcal{L}}_K$ loss was extremely noisy and therefore, no early stopping was performed. Nonetheless, other strategies could have been used such as stopping training when the discrepancy between the observed distribution $p(\y)$ and the regenerated one $\int q_{\phi}(\y|\x)q_{\theta}(\x)$ did not improve. $\mathcal{L}^{\text{ELBO}}$ and $\mathcal{L}^{\text{IW}}_K$ need more epochs to converge, likely due to the training of two networks simultaneously. When using these losses, training was done over 300 epochs over the whole observed dataset with $10\%$ of the data held out to stop training if the losses did not improve for 10 epochs.

\begin{table}[H]
\centering
\begin{tabular}{ll}
\hline \hline
     \multicolumn{2}{l}{\hspace{1.9cm} Architecture} \\ \hline
     Network architecture      & UMNN-MAF             \\ 
N$^\circ$integ. steps      & 20             \\ 
Embedding network       &         3 $\times$ 75  (MADE)   \\ 
Integrand network &  3 $\times$ 75 (MLP)           \\ 
N$^\circ$flows             &   6           \\ 
Embedding Size      &        10      \\ 
Batch size          &       128       \\
Optimizer          &        Adam      \\ 
Weight decay  &       0.0       \\ 
Learning rate   & $10^{-4}$             \\ \hline  \hline 
\end{tabular}
\caption{\label{tab:prior_params}Hyperparameters used to train and model $q_{\theta}(\x)$ and $q_{\psi}(\x|\y)$.}
\end{table}

\section{\label{appendix:M_1_EB}N=1 Empirical Bayes}

\begin{figure}[H]
\centering
\begin{subfigure}{.16\textwidth}
  \centering
  \includegraphics[width=1.\linewidth]{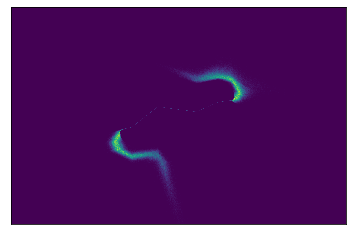}
\end{subfigure}
\begin{subfigure}{.16\textwidth}
  \centering
  \includegraphics[width=1.\linewidth]{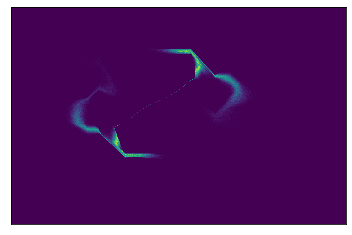}  
\end{subfigure}\\
\begin{subfigure}{.16\textwidth}
  \centering
  \includegraphics[width=1.\linewidth]{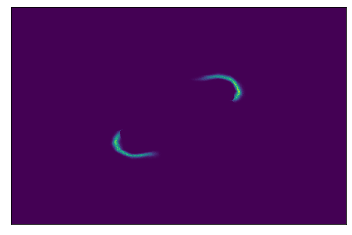}
\end{subfigure}
\begin{subfigure}{.16\textwidth}
  \centering
  \includegraphics[width=1.\linewidth]{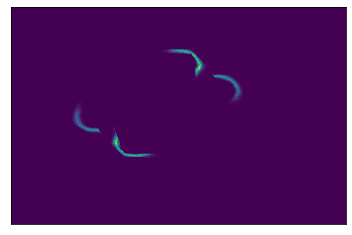}
\end{subfigure}\\
\begin{subfigure}{.16\textwidth}
  \centering
  \includegraphics[width=1.\linewidth]{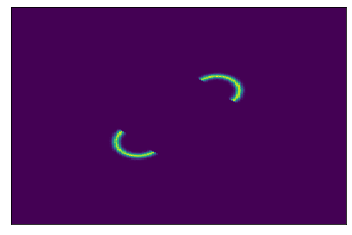}
    \caption{N=1.}
\end{subfigure}
\begin{subfigure}{.16\textwidth}
  \centering
  \includegraphics[width=1.\linewidth]{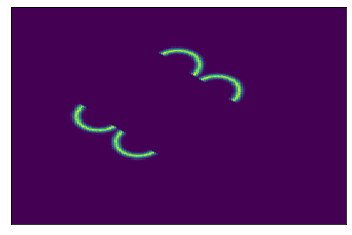}
    \caption{N=2.}
\end{subfigure}
\caption{\label{N=1_EB}Empirical Bayes with only $N=1$ or $N=2$ observations. (\textbf{Top row}) Learned (prior) distributions over source data. (\textbf{Middle row}) Learned distributions weighted by the likelihood approximated with the surrogate model. (\textbf{Bottom row}) Set of source data that may have generated the observation(s). Even with few observations, the method learns good posterior distributions.}
\end{figure}

Throughout the paper, the prior has been learned from the data given a large number of observations as it is often the case in the Empirical Bayes literature. Interestingly, Figure \ref{N=1_EB}  shows that even with a single (or two) observation(s), the method is able to learn the set of source data that may have generated the observation(s). 
When the number of observations is low, we observed that UMMN-MAFs tend to degenerate and concentrate all their masses to single points. Therefore, for this experiment, we used coupling layers that act as regularizers and do not collapse. We aim at studying the regularization introduced by bijective neural networks and how this may affect the learning of source data in the Neural Empirical Bayes framework in future work.

In this experiment, we used the $\mathcal{L}_{K}$ loss with $K=1024$. The distribution $q_{\theta}(\x)$ was modeled by 3 coupling layers where the scaling and translation networks are MLPs of 3 layers of 16 hidden units with ReLu activation.

\section{\label{appendix:MLP_as_prior} Empirical Bayes with simple models}

While the need to evaluate the density of new data points under the source model with $\mathcal{L}^\text{ELBO}$ and $\mathcal{L}^\text{IW}$ heavily restricts the model architectures that can be used to model $q_{\theta}(\x)$, $\mathcal{L}_\text{K}$ and $\hat{\mathcal{L}}_K$ allow to use any generative model mapping some noise $\z \in \mathbb{R}^n$ to $\x \in \mathbb{R}^d$. 

Normalizing flows have been consistently used in this paper to model $q_{\theta}(\x)$. Wile these models may in themselves act as a good inductive bias for continuous and smooth source distributions, we show here that simple MLPs can also learn good source distributions. This experiment is particularly useful as it shows that $\mathcal{L}^\text{K}$ and $\hat{\mathcal{L}}_K$ allow to use a broader class of model architectures than $\mathcal{L}^\text{ELBO}$ and $\mathcal{L}^\text{IW}$. This opens interesting research directions where useful inductive bias can be embedded in the source model. For example CNNs and RNNS can be used for image and time series analysis. 

In this experiment, we model $q_{\theta}(\x)$ with a $3-$layer MLPs with 100 units per layer and ReLU activations. We optimize $\theta$ with the same hyperparameters described in Appendix~\ref{appendix:hyper_params}. For a fixed GPU memory, the usage of simpler and lighter models allows to use higher values of $K$. In this experiment, we use $K=2^{10}$ and $K=2^{12}$.

\begin{table*}[h]
\centering
\setlength{\tabcolsep}{3pt}
\renewcommand{\arraystretch}{1.5}
\scriptsize
\begin{tabular}{l|ll|ll}
\hline\hline
\multirow{2}{*}{Simulator}                                   & \multicolumn{2}{c|}{$\y$-space}                                        &
\multicolumn{2}{c}{$\x$-space}  \\ 
                                                     & $\mathcal{L}_{1024}$& $\mathcal{L}_{4096}$ & $\mathcal{L}_{1024}$ & $\mathcal{L}_{4096}$  \\ \hline
SLCP  &   $0.55_{\pm 0.01}$ & $0.52_{\pm 0.01}$  &     $0.94_{\pm 0.01}$  &    $0.92_{\pm 0.01}$  \\ 
Two-moons &  $0.53_{\pm 0.02}$  &          $0.52_{\pm 0.01}$  & $0.68_{\pm 0.04}$ & $0.62_{\pm 0.05}$\\ 
\begin{tabular}[c]{@{}l@{}}IK\end{tabular} & $0.66_{\pm 0.03}$  & $0.58_{\pm 0.02}$     &    $0.92_{\pm 0.01}$              &    $0.90_{\pm 0.02}$               \\ \hline\hline
\end{tabular}
\caption{\label{tab:ROAC_AUC_appendix_table}Source estimation for the benchmark problems.
ROC AUC between $q_{\theta}(\x)$ and $p(\x)$ ($\x$-space), and between the observed distribution $p(\mathbf{y})$ and the regenerated distribution $\int p(\mathbf{y}|\mathbf{x})q_{\theta}(\mathbf{x})d\mathbf{x}$ ($\y$-space). }
\end{table*}

Table \ref{tab:ROAC_AUC_appendix_table} reports the discrepancy between the corrupted  data  from  the  identified  source  distributions and the ground truth distribution of noise-corrupted observations ($\y$-space). It shows that simple architectures allow to learn a source distribution that can closely reproduce the observed distribution. The ROC AUC between the source distribution $q_{\theta}(\x)$ and the ground truth distribution $p(\x)$ shows that the source distribution learned on the two-moons problem is close to the ground truth. For the other problems, useful inductive bias should be introduced to constrain the solution space.

\section{\label{appendix:symmetric_UMNN_MAF}Symmetric UMNN-MAF}

UMNN-MAF are autoregressive architectures such that:

\begin{equation}
    \x = \mathbf{G}(\z) = [g^1(z_1), ..., g^d(\z_{1:d})],
\end{equation}
where each $g^i(\cdot)$ is a bijective scalar function such that:

\begin{equation}
    g^i(\z_{1:i}) = \int_0^{z_i}f^i(t, \mathbf{h}^i(\z_{1:i-1}))dt + \beta^i(\mathbf{h}^i(\z_{1:i-1})),
\end{equation}
where $\mathbf{h}^i(\cdot) : \mathbb{R}^{i-1} \xrightarrow{} \mathbb{R}^q$ is a $q$-dimensional neural embedding of the variables $\z_{1:i-1}$, $f^i(\cdot) \in \mathbb{R}^+$ and $\beta^i(\cdot)$ is a scalar function.

In order to make the distribution $q_{\theta}(\x)$ one-to-one symmetric, i.e. $q_\theta(\left[x_1, ..., x_d\right]) = q_\theta(\left[\pm x_1, ..., \pm x_d\right]$), it is sufficient that (i) the distribution $p(\z)$ is one-to-one symmetric, (ii) $\beta^i(\cdot)$ is set to 0 and, (iii) the integrand function is such that $f^i(t, \mathbf{h}^i(x_1, ..., x_{i-1})) = f^i(\pm t, \mathbf{h}^i(\pm x_1, ..., \pm x_{i-1}))$. The condition (iii) is enforced by taking the absolute value of the input variables in the first layer of the integrand and embedding networks.

Then, $q_\theta(\left[x_1, ..., x_d\right]) = q_\theta(\left[\pm x_1, ..., \pm x_d\right]$.
\begin{proof}

First note that if conditions (ii) and (iii) are met:
\begin{equation}
    \label{eq:proof_symmetry}
    x^i = g^i(\pm z_1, ..., \pm z_{i-1}, z_i) \Leftrightarrow g^i(\pm z_1, ..., \pm z_{i-1}, -z_i) = -x^i 
\end{equation}
and 
\begin{equation}
    \label{eq:proof_symmetry_jacobian}
    |\det J_{g^i(\pm z_1, ..., \pm z_{i-1}, z_i)}| = |\det J_{g^i(\pm z_1, ..., \pm z_{i-1}, -z_i)}| = f^i(\pm z_i, \mathbf{h}^i(\pm z_1, ..., \pm z_{i-1})),
\end{equation}
where $J_{g^i(\pm z_1, ..., \pm z_i)}$ is the Jacobian of $g^i(\cdot)$ with respect to $z_i$.

It follows that:
    \begin{subequations}
        \begin{align}
           \label{eq:Proof_symmetric_density_a}
           q_\theta(\left[x_1, ..., x_d\right]) 
           &= p(z_1, ..., z_d)|\det J_{\G(\z)}|^{-1}\\  
           \label{eq:Proof_symmetric_density_b}
           &= p(z_1, ..., z_d)\prod_{i=1}^{d} f^i(z_i, \mathbf{h}^i(z_1, ..., z_{i-1}))^{-1}\\
           \label{eq:Proof_symmetric_density_c}
           &= p(\pm z_1, ..., \pm z_d)\prod_{i=1}^{d} f^i(\pm z_i, \mathbf{h}^i(\pm z_1, ..., \pm z_{i-1}))^{-1}\\
           \label{eq:Proof_symmetric_density_d}
           &= p(\pm z_1, ..., \pm z_d)|\det J_{\G(\pm z_1, ..., \pm z_d)}|^{-1}\\
           \label{eq:Proof_symmetric_density_e}
           &= q_\theta(\left[\pm x_1, ..., \pm x_d\right].
        \end{align}
    \end{subequations}
     Eq.~\ref{eq:Proof_symmetric_density_a} is a direct application of the change of variable theorem while Eq.~\ref{eq:Proof_symmetric_density_b} is obtained by definition. Conditions (i) and (iii) allow us to write Eq.~\ref{eq:Proof_symmetric_density_b} as Eq.~\ref{eq:Proof_symmetric_density_c}. The equalities in   Eq.~\ref{eq:proof_symmetry_jacobian} yields Eq.~\ref{eq:Proof_symmetric_density_d}. and finally, the last equation is obtained from Eq.~\ref{eq:Proof_symmetric_density_d} and Eq.~\ref{eq:proof_symmetry}.
\end{proof}

\section{\label{appendix:sbc} Simulation-Based
Calibration}

In order to perform simulation-based calibration, we repeatedly i) sample $\x^*$ from $p(\x)$, ii) generate $\y^*$ by running the simulator conditioned on $\x^*$, iii) perform rejection sampling in order to empirically approximate $p(\x|\y^*)$, and iv) store for each dimension $i$ in which quantile of $p(x_i|y_i^*)$, $x_i^*$ fall.

For each dimension, it is expected that $x\%$ of the parameters belong to the $x\%$ quantile of $p(x_i|y_i)$ and this can be assessed qualitatively by plotting the fraction of events per quantile as in figures~\ref{fig:KS_calibration} and~\ref{Calibration_test_herwig}. To perform a quantitative assessment, one can for example, compute the maximum absolute difference between the fraction of events that fall within a quantile and the value of that quantile, or in order words, report the Kolmogorov–Smirnov (KS) test between the empirical cumulative distribution function (blue lines on Figure~\ref{fig:KS_calibration} or Figure~\ref{Calibration_test_herwig})  and the expected cumulative distribution function (black line on Figure~\ref{fig:KS_calibration} or Figure~\ref{Calibration_test_herwig}). We report those quantities in Table~\ref{table:KS_test} for the different estimators.

The strength of this approach is that it allows to get insights about the learned posterior distribution without access to a ground truth. For example, if the model tends to assign more than $x\%$ of the parameters to the $x\%$ quantile, the model is overconfident. On the other hand, if it tends to assign less than $x\%$ of the parameters to the $x\%$ quantile, the model is underconfident.

In terms of weaknesses, the described approach only independently evaluate the 1D marginal distributions rather than the full-space distribution. In higher dimension, quantiles can be extended to contours but these might not be easily computable. Moreover, the approach is a necessary, but not sufficient, condition for well-calibrated posterior distribution. For example, by design $p(\x)$ would pass the calibration test.

\begin{table}[H]
\centering
\setlength{\tabcolsep}{2pt}
\renewcommand{\arraystretch}{1.5}
\scriptsize
\begin{tabular}{llc}
\hline \hline
 Simulator & Estimator & KS \\ \hline
\multirow{3}{*}{SLCP}  & $\mathcal{L}^{\text{ELBO}}$ & $0.37_{\pm0.01}$    \\ 
                       & $\mathcal{L}^{\text{IW}}_{128}$ & $0.15_{\pm0.02}$   \\ 
                       & $\mathcal{L}_{1024}$ & \bm{$0.14_{\pm0.01}$}  \\ \hline
\multirow{3}{*}{\begin{tabular}[c]{@{}l@{}}Two-moons\end{tabular}} 
                       & $\mathcal{L}^{\text{ELBO}}$     &   $0.40_{\pm0.01}$       \\ 
                       & $\mathcal{L}^{\text{IW}}_{128}$ &   $0.45_{\pm0.01}$      \\  
                       & $\mathcal{L}_{1024}$     &     \bm{$0.10_{\pm0.01}$}    \\ \hline
\multirow{3}{*}{\begin{tabular}[c]{@{}l@{}}IK\end{tabular}}                                                    & $\mathcal{L}^{\text{ELBO}}$     &  $0.65_{\pm0.04}$        \\ 
            & $\mathcal{L}^{\text{IW}}_{128}$     &  $0.47_{\pm0.05}$        \\ 
            & $\mathcal{L}_{1024}$     &  \bm{$0.09_{\pm0.02}$}       \\ \hline \hline
\end{tabular}
\caption{\label{table:KS_test}Calibration test from 1000 posterior estimates obtained with rejection sampling for $\mathcal{L}_{1024}$, importance sampling for $\mathcal{L}^{\text{IW}}_{128}$ and directly from the recognition network $q_{\psi}(\x|\y)$ for $\mathcal{L}^{\text{ELBO}}$.
\textit{As opposed to $\mathcal{L}_{1024}$, the posterior distributions for $\mathcal{L}^{\text{IW}}_{128}$ and $\mathcal{L}^{\text{ELBO}}$ are not consistently correctly calibrated.}}
\end{table}

\section{ 
\label{appendix:phys_simulator}  Collider Physics Simulation}

The simulated physics dataset, made publically available by~\cite{Andreassen2019OmniFoldAM}, targets conditions similar to those produced by the proton-proton collisions  at $\sqrt{s} = 14$ TeV at the Large Hadron Collider~\citep{Evans_2008}.  For surrogate training, source distributions of jets from collisions producing $Z$ bosons recoiling off of jets are modeled with the the Monte Carlo simulator Pythia 8.243~\citep{SJOSTRAND2015159} with Tune 26~\citep{ATL-PHYS-PUB-2014-021}. For learning the source distribution with NEB, an alternative simulation of the source distribution of jets from collisions producing $Z$ bosons recoiling off of jets is performed with Herwig 7.1.5~\citep{B_hr_2008, Bahr1999Herwig2R} with default tune. The Delphes simulator~\citep{de_Favereau_2014} is used to model the impact of detector effects on particle measurements using a parameterized detector smearing that models the smearing effects in the ATLAS~\citep{ATLAS_2008} or CMS~\citep{CMS_2008} experiments.

\end{document}